\documentclass[10pt,journal,compsoc]{IEEEtran}

\ifCLASSOPTIONcompsoc
  \usepackage[nocompress]{cite}
\else
  \usepackage{cite}
\fi
\usepackage[T1]{fontenc}
\usepackage{graphicx}
\usepackage{amsmath}
\usepackage{url}
\usepackage{amssymb}
\usepackage{booktabs}
\usepackage{tikz}
\usepackage{etoolbox}
\usepackage{ragged2e}
\usepackage{enumitem}
\usepackage{caption}
\usepackage{multirow}
\newcommand*{\circled}[1]{\lower.7ex\hbox{\tikz\draw (0pt, 0pt)%
    circle (.5em) node {\makebox[1em][c]{\small #1}};}}
\robustify{\circled}
\usepackage[colorlinks,linkcolor=blue]{hyperref}

\usepackage{algorithm}
\usepackage{algorithmic}

\ifCLASSINFOpdf
\else
\fi

\begin{document}
\title{Learning Heterogeneous Mixture of Scene Experts for
Large-scale Neural Radiance Fields}

\author{Zhenxing Mi,~Ping~Yin,~Xue~Xiao,~and~Dan~Xu\IEEEauthorrefmark{1}~\IEEEmembership{Member,~IEEE}
\IEEEcompsocitemizethanks{\IEEEcompsocthanksitem Zhenxing Mi and Dan Xu are with the Department
of Computer Science and Engineering, The Hong Kong University of Science and Technology, Hong Kong SAR.
\protect
E-mail: zmiaa@connect.ust.hk, danxu@cse.ust.hk

\IEEEcompsocthanksitem Ping Yin and Xue Xiao are with Inspur Cloud Information Technology Co, Ltd. E-mail: yinping@inspur.com, xiaoxue@inspur.com

\IEEEcompsocthanksitem\IEEEauthorrefmark{1}Corresponding author

}%
}

\markboth{Journal of \LaTeX\ Class Files,~Vol.~14, No.~8, August~2015}%
{Shell \MakeLowercase{\textit{et al.}}: Bare Demo of IEEEtran.cls for Computer Society Journals}

\IEEEtitleabstractindextext{%
\begin{abstract}

Recent Neural Radiance Field (NeRF) methods on large-scale scenes have demonstrated promising results and underlined the importance of scene decomposition for scalable NeRFs. Although these methods achieved reasonable scalability, there are several critical problems remaining unexplored in the existing large-scale NeRF modeling methods, i.e., learnable decomposition, modeling scene heterogeneity, and modeling efficiency. In this paper, we introduce Switch-NeRF++, a Heterogeneous Mixture of Hash Experts (HMoHE) network that addresses these challenges within a unified framework. Our framework is a highly scalable NeRF that learns heterogeneous decomposition and heterogeneous Neural Radiance Fields efficiently for large-scale scenes in an end-to-end manner. In our framework, a gating network learns to decompose scenes into partitions and allocates 3D points to specialized NeRF experts. This gating network is co-optimized with the experts by our proposed Sparsely Gated Mixture of Experts (MoE) NeRF framework. Our network architecture incorporates a hash-based gating network and distinct heterogeneous hash experts. The hash-based gating efficiently learns the decomposition of the large-scale scene. The distinct heterogeneous hash experts consist of hash grids of different resolution ranges. This enables effective learning of the heterogeneous representation of different decomposed scene parts within large-scale complex scenes. These design choices make our framework an end-to-end and highly scalable NeRF solution for real-world large-scale scene modeling to achieve both quality and efficiency. We evaluate our accuracy and scalability on existing large-scale NeRF datasets. Additionally, we also introduce a new dataset with very large-scale scenes ($>6.5km^2$) from UrbanBIS. Extensive experiments demonstrate that our approach can be easily scaled to various large-scale scenes and achieve state-of-the-art scene rendering accuracy. Furthermore, our method exhibits significant efficiency gains, with an 8x acceleration in training and a 16x acceleration in rendering compared to the best-performing competitor Switch-NeRF. The codes and trained models will be released in \href{https://github.com/MiZhenxing/Switch-NeRF}{https://github.com/MiZhenxing/Switch-NeRF}.

\end{abstract}

}

\maketitle

\IEEEdisplaynontitleabstractindextext

\IEEEpeerreviewmaketitle

\IEEEraisesectionheading{\section{Introduction}\label{sec:introduction}}

Neural Radiance Fields (NeRF)~\cite{mildenhall2020nerf} have shown the powerfulness for high-fidelity 3D scene modeling. Recent methods~\cite{Turki_2022_CVPR, Tancik_2022_CVPR, mi2023switchnerf} have scaled NeRF to large-scale scenes with thousands of high-resolution images. The scalability of them essentially comes from the decomposition of the 3D scenes. Several sub-networks handle different partitions of a large scene. Mega-NeRF~\cite{Turki_2022_CVPR} and Block-NeRF~\cite{Tancik_2022_CVPR} both incorporate hand-crafted decomposition rules for different scenes. They decompose 3D scenes by clustering based on 3D physical distances or street blocks.

Despite the promising performances achieved by current large-scale NeRF methods, several critical challenges remain unexplored. \textbf{\emph{Firstly}}, designing a universal decomposition rule is extremely challenging for different large-scale scenes. The real-world large-scale scenes tend to have very different and complex scene structures. Hand-crafted rules will inevitably bring adaptation issues for distinct scenarios. 
Moreover, the non-learnable nature of hand-crafted decomposition restricts the network's ability to optimize the scene decomposition and radiance fields together within an end-to-end framework. To handle this problem, we need a learnable decomposition method to automatically learn different decompositions for different scenes, and we can optimize the decomposition together with the scene representation.
\textbf{\emph{Secondly}}, a large-scale scene is typically irregular and complex, with different scene parts exhibiting diverse appearance and geometry distributions, which requires heterogeneous modeling capabilities to produce high-quality scene representation.
However, current large-scale NeRF methods often employ identical networks, such as Multilayer Perceptrons (MLPs), for different parts of the scene, lacking explicit designs for heterogeneous modeling. \textbf{\emph{Thirdly}}, the long training time and high GPU memory usage of existing methods are a severe bottleneck that limits their scalability when dealing with much larger scenes. Although several methods such as Instant-NGP~\cite{mueller2022instant} have been proposed to enhance the NeRF representation and accelerate the NeRF training on small- and large-scale scenes~\cite{SunSC22, yu_and_fridovichkeil2021plenoxels, Chen2022ECCV, xu2023gridguided}, they still attempt to fit an entire scene into a single global network, overlooking the crucial aspect of heterogeneous scene representation from distinct scene decompositions. Consequently, these methods suffer from clear limitations in scalability and representation effectiveness, particularly when dealing with very large scenes.

\begin{figure*}[t]
\begin{center}
   \includegraphics[width=\textwidth]{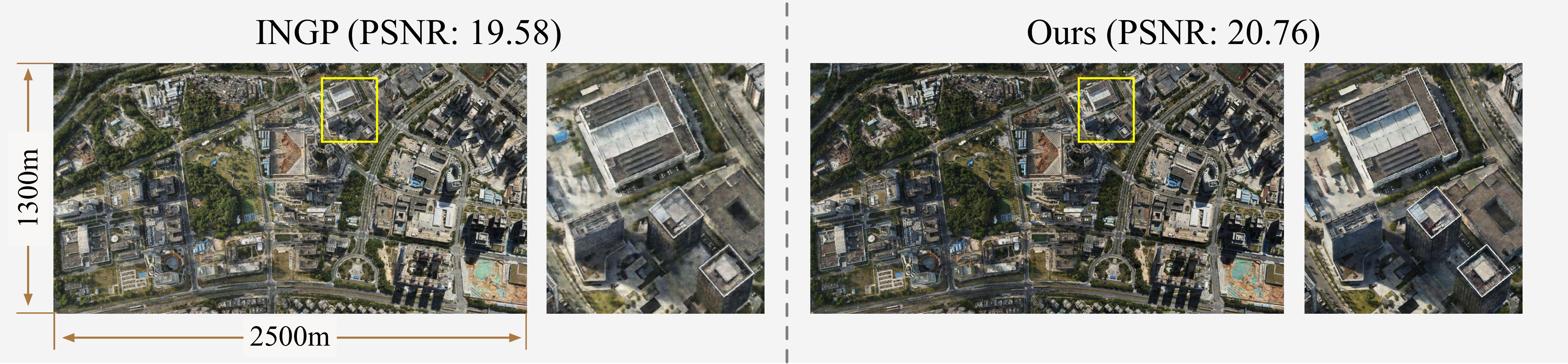}
\end{center}
\vspace{-10pt}

   \caption{The bird's-eye view of a scene in UrbanBIS~\cite{UrbanBISbib}, rendered using INGP~\cite{mueller2022instant} and our Switch-NeRF++, showcasing the impressive scalability of our approach. This scene contains $13k$ high-quality images, capturing an urban region of around $6.5km^2$. Experiments on this scene demonstrate the superiority of our method in scaling to large-scale urban scenes with exceptional rendering quality.}

\vspace{-15pt}

\label{fig:teaser}
\end{figure*}

\par In this paper, to simultaneously tackle these critical challenges, we propose a novel and highly scalable Heterogeneous Mixture of Hash Experts framework for NeRF, coined as Switch-NeRF++. The fundamental designs of Switch-NeRF++ are fourfold. 1) An end-to-end framework to learn scene decomposition in a trainable manner, together with the scene representation. 2) An efficient hash-based gating network to learn scene decompositions. 3) a heterogeneous modeling of scene decomposition and each decomposed scene part. 4) a highly scalable and efficient learning framework that can handle extremely large-scale scenes. With these key designs, Switch-NeRF++ can handle large-scale complex scenes and ensure high modeling performance in terms of scalability, representation quality, and efficiency.
More specifically, in Switch-NeRF++, we design:

\noindent \textbf{A sparse framework for unified optimization of scene decomposition and neural radiance fields.} We introduce an end-to-end sparse framework that unifies optimization of scene decomposition and Neural Radiance Fields by a learnable gating mechanism. It employs a trainable gating network to dynamically select and sparsely activate specialized experts for individual 3D points. The whole network is trained end-to-end without heuristic supervision and priors of the 3D scene shape or the distribution of scene images. This enables scalable and generic modeling of large-scale scenes.
A critical challenge of training our network end-to-end lies in the discrete selection operation of experts. This operation is not differentiable, and the gradients cannot be back-propagated through it into the gating network. Inspired by Sparsely-Gated Mixture-of-Experts (MoE)~\cite{DBLP:conf/iclr/ShazeerMMDLHD17}, we address this problem by multiplying gate values to the expert outputs, adding a connection between the gating network and experts. For the scene representation, distinct scene regions are handled by different NeRF experts. The gating network routes 3D points to relevant experts based on gate values. A unified output head is designed to unify the prediction of multiple NeRF experts. This head imposes feature alignment among the implicit features of NeRF experts, resolving inconsistencies among different partitions.

\noindent \textbf{Heterogeneous Mixture of Hash Experts (HMoHE) to model heterogeneous scene representations.} Previous large-scale NeRF methods typically use the same network structure for different experts. However, different parts of the large-scale scenes have heterogeneous and diverse distributions, which need heterogeneous representations. In our framework, the learned decomposition of the entire scene is represented by a mixture of heterogeneous hash experts (HMoHE). Each decomposed scene part is handled by a distinct expert network. To ensure the diversity and flexibility of representation, we design the hash experts as heterogeneous. We make the experts exploit diverse resolution ranges, allowing our network to select a suitable resolution range for each scene part.

\noindent \textbf{A hash-based gating network to effectively learn the scene decomposition.} The gating network dispatches 3D points into different hash-encoding experts, enabling each expert to specialize in representing a distinct decomposed partition of the scene and to achieve a stronger representation. The hash-based gating network is more efficient than MLP-based gating~\cite{fedus2022switch} to make the training of gating align well with our hash-encoding experts. The hash-based gating network and HMoHE together achieve accurate and efficient heterogeneous scene representations for large-scale scenes.

\noindent \textbf{Efficient implementation of hash-based gating and dispatching.} To implement HMoHE more efficiently, we seamlessly integrate the dispatching of 3D points into the CUDA implementation of hash. We merge the hashing and dispatching of points together and avoid extra allocation of memory. This ensures that the point dispatching does not significantly impact the training efficiency. 

Moreover, apart from existing challenging large-scale datasets, we adapt a very large scene from UrbanBIS~\cite{UrbanBISbib}. This scene enables us to assess the scalability of our approach more comprehensively. Our results show that Switch-NeRF++ achieves remarkably higher rendering quality than existing best-performing methods. These results clearly verify the effectiveness and high scalability of our method for large-scale scenes.

This paper significantly extends our previous ICLR version, i.e., Switch-NeRF \cite{mi2023switchnerf}. Specifically, this work introduces a brand new design of the Heterogeneous Mixture of Hash framework. We redesign the gating network and experts in Switch-NeRF, exploiting highly efficient hash networks. We propose to learn heterogeneous decomposition and representation with the hash gating network and heterogeneous experts. For the point dispatching, we integrate it into the hash operation. We also adapt a new large-scale scene for better evaluation. Compared to Switch-NeRF, our new framework effectively models the heterogeneous representation of large-scale scenes and significantly improves the accuracy and efficiency. The proposed method achieves an 8x acceleration in training and a 16x acceleration in rendering compared to Switch-NeRF.

Our contributions are summarized below:

\begin{itemize}[leftmargin=*]

    \item We introduce a novel framework for large-scale neural radiance fields (NeRF), which integrates learnable scene decomposition, heterogeneous representation learning, and efficient modeling into a unified end-to-end architecture. Unlike prior methods, our framework does not rely on hand-crafted heuristic decomposition and is generally applicable to large-scale scene modeling.

    \item As the core of our framework, we design a hash-based gating network that learns to decompose scenes automatically and efficiently. We introduce Heterogeneous Mixture of Hash Experts (HMoHE) to enable learning scale-aware distributions. The gating and expert modules are jointly optimized to produce highly scalable and heterogeneous NeRF representations.

    \item We present an efficient implementation of the framework to make it a highly scalable and efficient learning framework that can handle extremely large-scale scenes. We adapt a new dataset for benchmarking highly scalable NeRF scene modeling. Our NeRF framework achieves significant improvement in terms of scalability, accuracy, and efficiency compared to the best-performing methods such as Switch-NeRF.

\end{itemize}

\begin{figure*}[t]
\centering
    \includegraphics[width=\textwidth]{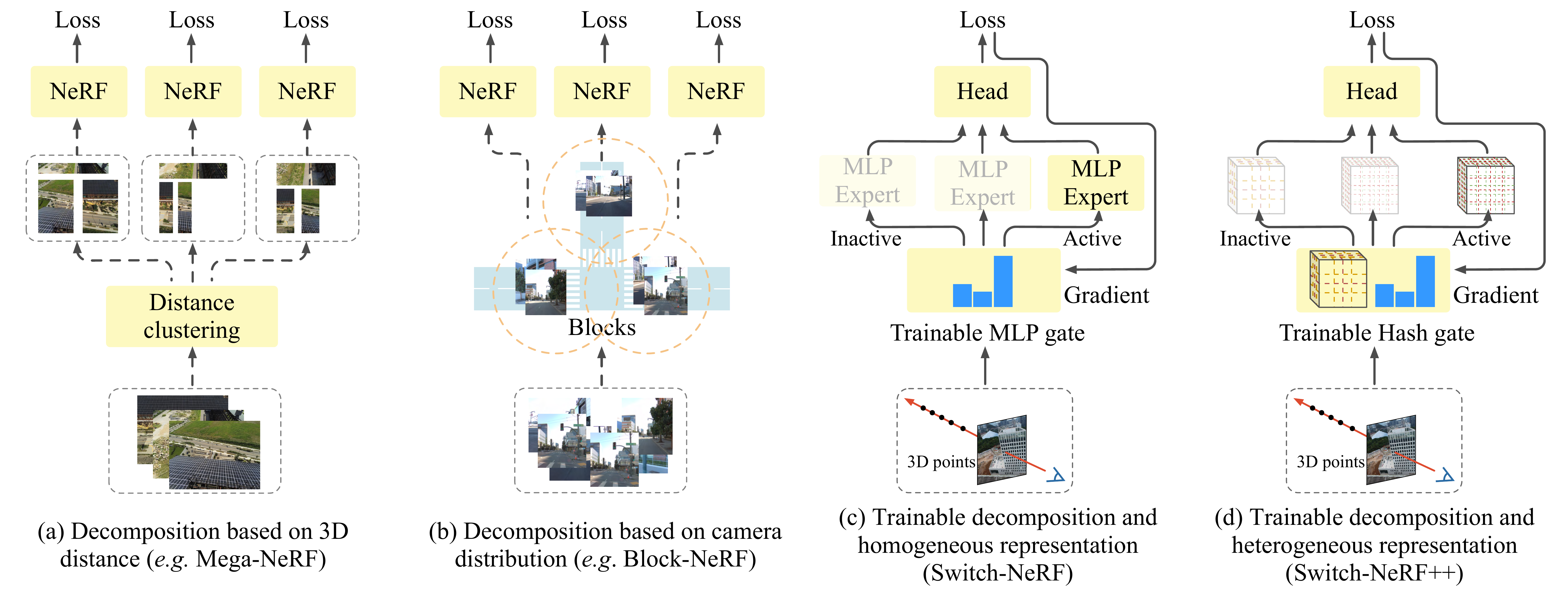}

  \caption{Large-scale NeRFs with different decomposition methods. Dotted lines indicate non-differentiable operations, while solid lines represent differentiable operations that can back-propagate gradients in training. Mega-NeRF \cite{Turki_2022_CVPR} divides the scene by clustering image pixels by 3D distances. Block-NeRF \cite{Tancik_2022_CVPR} divides image sets according to street blocks, which is actually based on camera distributions. In both kinds of methods, the decompositions are hand-crafted heuristics rules and not trainable. They require separate training for their sub-networks. Switch-NeRF introduces a unified framework where scene decomposition is learned automatically via a gating network, enabling end-to-end training of the entire framework. However, Switch-NeRF designs homogeneous experts with the same structures, which learns homogeneous representations of the scene. In contrast, Switch-NeRF++ employs an efficient hash network as the gate network and designs heterogeneous hash experts to learn heterogeneous decomposition and representation of a large-scale scene.}
  \label{fig:archcompare}

\end{figure*}

\begin{figure*}[htbp]
  \centering
  \includegraphics[width=\linewidth]{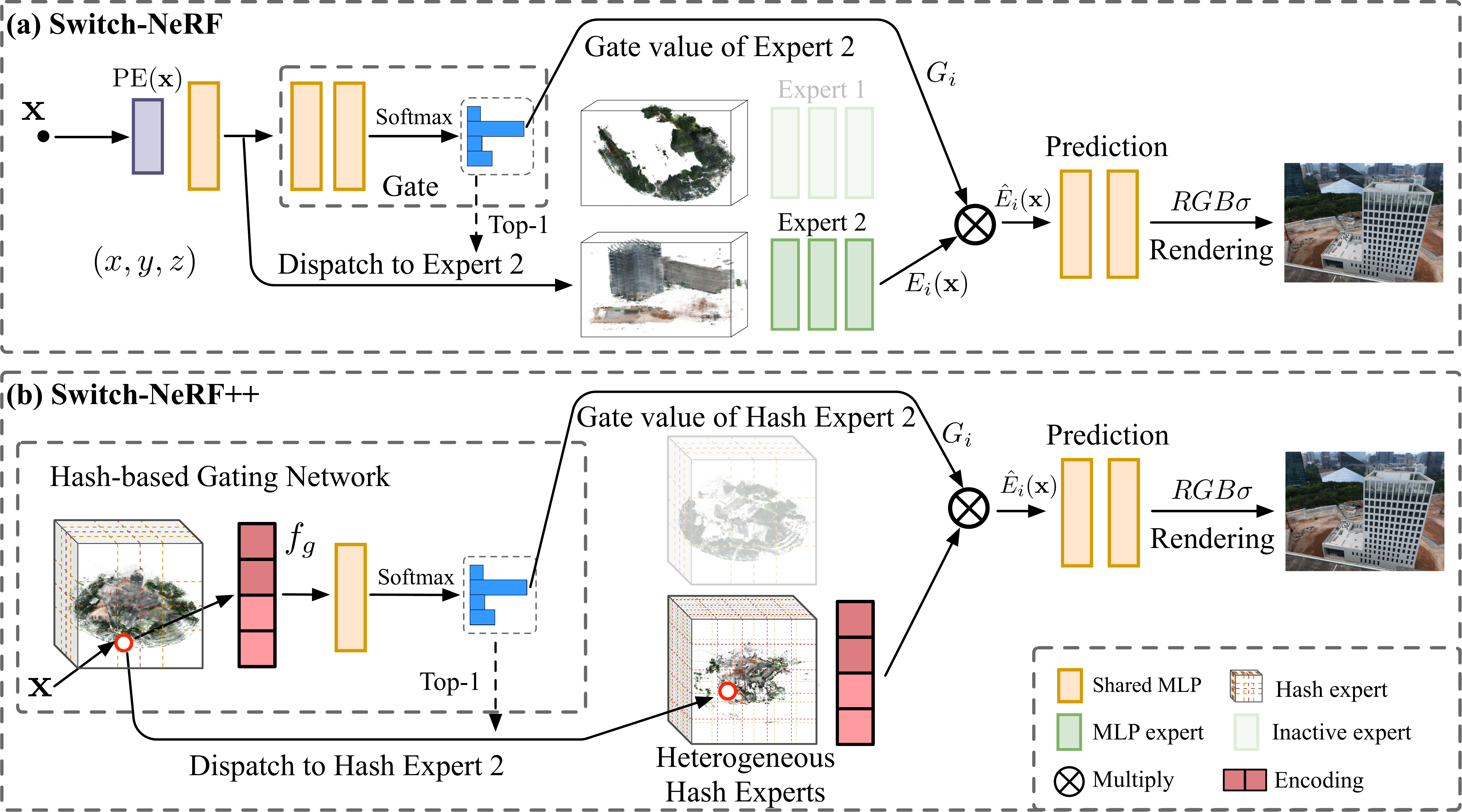}
  \caption{The framework overview of Switch-NeRF and Switch-NeRF++ for large-scale scene modeling. In Switch-NeRF++, A 3D scene point \textbf{x} is first fed into the gating network to obtain a feature encoding $f_g$. $f_g$ is then sent to a small MLP head to predict gating values. The Top-$1$ index $i$ of the gate values is used in 3D point dispatching that determines which hash expert should be selected for the input point. In this way, the entire scene can be decomposed into different partitions. Then $\textbf{x}$ is dispatched into its corresponding hash expert to encode its distinct distribution into a hash expert encoding $E_i(\textbf{x})$. 
  Then we multiply the gate value $G_i(\textbf{x})$ on $E_i(\textbf{x})$ and obtain the final hash expert output $\hat{E}_i(\textbf{x})$, which is further fed into a small MLP head to predict the point density $\sigma$ and point color $\textbf{c}$. The spherical harmonics encoding (\emph{i.e.}, $\mbox{SH}$) of direction $\textbf{d}$, and the appearance embedding (\emph{i.e.}, $\mbox{AE}$) are used for view-dependent prediction of color $\textbf{c}$. In Switch-NeRF, the gating network and the experts are all MLP-based networks, which learns homogeneous representations. In contrast, in Switch-NeRF++, we leverage the efficient Hash-based Gating Network to dispatch 3D points. Furthermore, we design heterogeneous hash experts that have different scale ranges to explicitly model the heterogeneous distribution of the large-scale scene.}
  \label{fig:framework}
  \vspace{-10pt}
\end{figure*}

\section{Related Work}

\noindent \textbf{3D reconstruction from multiple images.}
3D reconstruction from multiple images has been a long-standing research topic in computer vision. A classic pipeline contains several steps. The first step is to extract feature points from multiple images by methods such as SIFT \cite{lowe1999object} and match corresponding points across different images \cite{cech2010efficient, schonberger2017vote, sarlin2020superglue}.
Then the corresponding points are used in Structure-from-motion (SFM) methods \cite{snavely2006photo, crandall2011discrete, schonberger2016structure} to recover the camera poses. After recovering camera poses, Multi-view Stereo (MVS) methods can be applied on the images and cameras to recover the dense geometry of the scene, using optimization-based methods~\cite{furukawa2009accurate,schonberger2016pixelwise,xu2019multi} or deep learning-based methods~\cite{mi2022generalized,yao2018mvsnet, chen2019point, gu2020cascade, xu2020learning, mi2022generalized}.

\noindent \textbf{Neural Radiance Fields.} The Neural Radiance Fields method (NeRF) \cite{mildenhall2020nerf} is proposed to encode a 3D scene into an MLP network for novel view synthesis, with the multiple images and their cameras as input. The implicit framework of NeRF is very simple and effective, and has been widely applied to many 3D reconstruction tasks. 

Some NeRF variants apply NeRF to more challenging scenes. NeRF++~\cite{zhang2020nerf++} extends NeRF to unbounded scenes by separating a scene into a foreground and a background, and applying the inverted sphere parameterization on the background. NeRF in the wild~\cite{martin2021nerf} attaches an appearance embedding to each image. It conditions the NeRF network with these appearance embeddings to model the appearance variation in each image in the wild environment.
Mip-NeRF~\cite{barron2021mip} and Mip-NeRF 360~\cite{ barron2022mip} address sampling and aliasing of bounded and 360-degree scenes by designing a new scale-aware positional encoding.
DSNeRF~\cite{kangle2021dsnerf} uses sparse depth from colmap to regularize and speed up NeRF training.
SPARF~\cite{sparf2023} and CVT-xRF~\cite{zhong2024cvt} largely reduce the view number used in NeRF training and produce high-quality NeRF with sparse image inputs.

Although the original NeRF targets novel view synthesis, it is a flexible 3D representation encoding 3D geometry. Several methods have focus on improving the geometry quality of NeRF for accurate 3D surface reconstruction~\cite{wang2021neus,li2023neuralangelo}. NeuS~\cite{wang2021neus} represents a surface as a signed distance function (SDF) and learns an SDF field by volume rendering. Neuralangelo~\cite{li2023neuralangelo} recovers high-quality surface details by higher-order derivatives and coarse-to-fine optimization on the hash grids.

NeRF methods typically need camera parameters as input. These cameras are usually predicted by SFM methods such as Colmap~\cite{schoenberger2016sfm}. Several methods try to estimate camera parameters together with the neural radiance fields. NeRF{-}{-}~\cite{wang2021nerf} treats camera parameters as trainable model parameters and jointly trains cameras in NeRF training. BARF~\cite{lin2021barf} finds that using coarse-to-fine positional encoding in training helps the convergence of camera parameters.

The training and inference speed of NeRF is critical in real-world applications. Many methods have been proposed for fast training and real-time rendering of NeRF. PlenOctrees~\cite{yu2021plenoctrees}, FastNeRF~\cite{garbin2021fastnerf}, and Kilo-NeRF~\cite{Reiser_2021_ICCV} mainly speed up inference of NeRF. There are also methods to speed up both the training and rendering of NeRF, such as the DVGO~\cite{SunSC22}, Plenoxels~\cite{Fridovich_Keil_2022_CVPR}, Instant-NGP~\cite{mueller2022instant}, TensoRF~\cite{Chen2022ECCV}, and 3D Gaussian Splatting~\cite{kerbl3Dgaussians}. Instant-NGP uses a multi-resolution hash encoding to encode a scene into a set of hash maps by interpolating trainable hash features. TensoRF~\cite{Chen2022ECCV} represents a 3D radiance field using trainable 2D plane features and 1D feature vectors. Different from typical NeRF methods using volume rendering, 3D Gaussian Splatting (3DGS)~\cite{kerbl3Dgaussians} used 3D Gaussians to represent a 3D scene. It renders images by projection and $\alpha$-blending, which is very fast and supports high-resolution real-time rendering. 3DGS needs a point cloud from SfM or random initialization to initialize 3D Gaussians. 

The original NeRF needs to be trained on each scene and cannot generalize across different scenes. Pixel-NeRF~\cite{yu2021pixelnerf} proposes to include image features as an additional condition and trains the model on a set of different scenes. MVSNeRF~\cite{chen2021mvsnerf}, NerfingMVS~\cite{wei2021nerfingmvs}, and IBRNet~\cite{wang2021ibrnet} et. al. handle the unsupervised Multi-view Stereo problem by combining volume rendering and Multi-view Stereo together. These methods take single-view or multi-view image features as input and can generalize to different scenes.

The flexible NeRF representation has been applied to 3D generation~\cite{chan2022efficient, honglrm,dong2024interactive3d} and editing tasks~\cite{spinnerf, wang2024gscream}. EG3D~\cite{chan2022efficient} simplifies the NeRF network into triplane feature maps, which have regular shapes and are easy for generation. LRM~\cite{honglrm} proposes to use a large transformer to generate triplanes from a single image. 

The above-mentioned reconstruction methods typically work on small-scale scenes. Our method aims to scale NeRF methods to large-scale scenes and handle the problems of learnable decomposition, and heterogeneous, scalable, and efficient scene modeling for large-scale NeRF.

\noindent \textbf{Large-scale Neural Radiance Fields.} When applied to large-scale scenes, a critical problem of NeRF is that handling a large data volume requires a scalable framework and more network parameters. Recent large-scale NeRF methods mainly focus on decomposing a large-scale scene into several parts and handling each part by a sub-network. Mega-NeRF~\cite{Turki_2022_CVPR} and Block-NeRF~\cite{Tancik_2022_CVPR} successfully scale NeRF to urban-scale scenes sparsely by scene decomposition. Although achieving impressive accuracy, their scene decomposition is mainly conducted by heuristic hand-crafted rules. Mega-NeRF~\cite{Turki_2022_CVPR} places several centroids in a scene and divides the 3D points in the scene by 3D distances to these centroids. Block-NeRF~\cite{Tancik_2022_CVPR} divides the training images into several subsets according to street blocks. These two methods train the sub-networks separately. Therefore, they need composition strategies such as distance-based weights to merge the rendering results. Different from methods using handcrafted decomposition methods, Switch-NeRF~\cite{mi2023switchnerf} firstly proposes a large-scale NeRF framework that automatically learns scene decomposition together with the Neural Radiance Fields in an end-to-end manner. It designs a sparse network structure containing a gating network and several NeRF experts, based on the mechanism of Mixture of Experts (MoE)~\cite{DBLP:conf/iclr/ShazeerMMDLHD17}. It requires no prior knowledge of scene structures and decomposition rules. Our method extends from Switch-NeRF with brand-new designs of the hash-based network, heterogeneous experts, and efficient hash dispatching. The Mega-NeRF, Block-NeRF and our Switch-NeRF++ share a divide-and-conquer paradigm for large-scale scene reconstruction. However, we use fundamentally different decomposition methods and training strategies to scale the NeRF. As illustrated in Fig.~\ref{fig:archcompare}, our Switch-NeRF++ introduces a unified framework that incorporates the decomposition and scene representation into a single learning network. It not only learns the scalable and heterogeneous scene representation by sparse experts, but also learns the heterogeneous decomposition itself. With the hash-based network, heterogeneous experts and efficient hash dispatching, and unified heads, our method demonstrates strong abilities to learn an accurate and consistent large-scale NeRF on various large-scale scenes. In contrast, Mega-NeRF and Block-NeRF use handcrafted rules to decompose the scene. Each of its sub-networks is trained independently. This makes the decomposition not aware of the scene structure, and the learned representations are not consistent across scenes. Our method remarkably improves the scalability, efficiency, and accuracy of large-scale NeRFs compared to best-performing methods.

Some other methods~\cite{zhang2023efficient, xu2023gridguided} also use plane features and hash encodings on large-scale scenes. However, they still encode a large-scale scene into a global network, so their scalability is still limited. The Gaussian splatting~\cite{kerbl20233d} has inspired large-scale NeRFs such as the Hierarchical-3D-Gaussians~\cite{hierarchicalgaussians24}, CityGaussian~\cite{liu2024citygaussian}, DoGaussian~\cite{chen2024dogaussian}, VastGaussian~\cite{lin2024vastgaussian}, PyGS~\cite{wang2024pygs}, OccluGaussian~\cite{liu2025occlugaussian}, etc. due to its efficiency. The Hierarchical-3D-Gaussians~\cite{hierarchicalgaussians24} uses a divide-and-conquer method that trains very large scenes in independent chunks. It organizes the chunks into a hierarchy and provides an efficient Level-of-Detail (LOD) rendering. CityGaussian~\cite{liu2024citygaussian} and VastGaussian~\cite{lin2024vastgaussian} also use a divide-and-conquer method to manually divide the scene into smaller cells. 
In this paper, our method focuses on scaling to large-scale scenes by learning efficient scene decomposition and representations. Our method explores the better use of the multi-scale nature of grids to adaptively model the multi-scale nature of 3D scenes.

\noindent \textbf{Mixture of Experts (MoE)}. 
The methodology of Mixture of Experts has a long history for scaling the model size~\cite{jacobs1991adaptive, jordan1994hierarchical}. Recent applications of MoE in deep learning mainly follow the design of \cite{DBLP:conf/iclr/ShazeerMMDLHD17}. It replaces the feed-forward networks in a language model with Mixture-of-Experts layers. The experts are activated sparsely by a Top-$k$ gating network. Each sample will only be dispatched to $k$ experts while $k$ is typically set as 2 and is much smaller than the total number of experts. In order to avoid imbalanced training and training collapse, it regularizes gate values by a gate loss, which can make the training of different experts more balanced. The MoE network is extensively applied for Natural Language Processing (NLP) \cite{DBLP:conf/iclr/LepikhinLXCFHKS21, fedus2022switch} and Vision~\cite{DBLP:conf/nips/RiquelmePMNJPKH21,ye2023taskexpert} tasks to scale the capacity and capability of neural networks. GShard~\cite{DBLP:conf/iclr/LepikhinLXCFHKS21} standardize the dispatching operations by Einops \cite{rogozhnikov2021einops, DBLP:conf/iclr/LepikhinLXCFHKS21}. It supports local dispatching and parallel gating by designing a simpler balancing loss and defining the capacity factor for experts.
The value of $k$ in MoE gating directly influences the sparsity, efficiency, and accuracy of MoE. Top-$2$ typically can get good accuracy while maintaining sparsity~\cite{DBLP:conf/iclr/ShazeerMMDLHD17, DBLP:conf/iclr/LepikhinLXCFHKS21}.
Switch Transformer \cite{fedus2022switch} has investigated the training recipes of MoE in NLP tasks. It concludes that under a carefully designed training setting, Top-$2$ operation in MoE is not necessary. Top-$1$ in MoE can already get high accuracy while largely decreasing the cost of computation and communication for sample dispatching. Several MoE methods have investigated different gating mechanisms, such as Hash Routing~\cite{roller2021hash}, BASE~\cite{lewis2021base}, and Expert Choice Routing~\cite{zhou2022mixture}. 

Since MoE is typically used to scale up deep learning models, it is often trained on a large number of GPUs. This poses challenges to the efficient implementation of MoE and the design of a large-scale training system. Mesh-TensorFlow \cite{shazeer2018mesh} provides a multiple-node implementation of MoE based on TensorFlow. Deepspeed~\cite{DBLP:conf/icml/RajbhandariLYZA22} uses their efficient inference and training system to train MoE at scale and proposes a residual MoE design. Tutel \cite{hwang2022tutel} introduces optimized parallelism strategies and CUDA kernels to train large-scale MoE systems. The dispatching operation of Switch-NeRF is based on Tutel kernels, while Switch-NeRF designs the full dispatching operation to improve the model accuracy. In Switch-NeRF++, we innovatively merge the dispatching into the hash operation and largely decrease the computation and memory cost of our MoE system.

\begin{figure*}[htbp]
\centering
    \includegraphics[width=\linewidth]{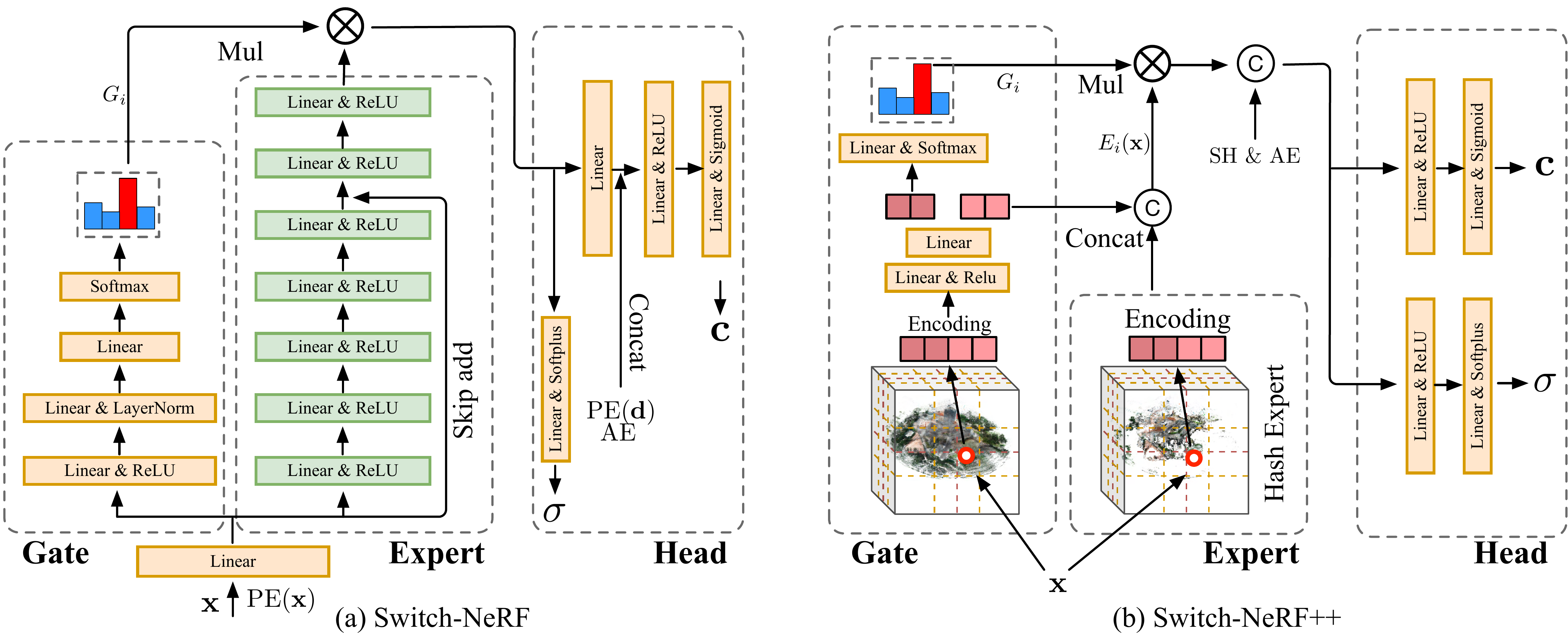}

  \caption{(a) A detailed illustration of the MLP-based gating network with an MLP expert in Switch-NeRF. (b) A detailed illustration of the hash-based gating network and a hash expert in Switch-NeRF++. An input \textbf{x} first goes through the gating network. Then it goes through a selected expert and the small prediction heads.}

  \label{fig:gateexpert}
\end{figure*}

\section{Switch-NeRF++}
Our Switch-NeRF++ learns to encode a large-scale scene into heterogeneous, efficient, and scalable Mixture of Hash Experts. Fig.~\ref{fig:framework} is the overview of the framework in which Fig.~\ref{fig:framework}a and Fig.~\ref{fig:framework}b show the structure of Switch-NeRF and Switch-NeRF++, respectively. Let $\textbf{x}$ denote a 3D point and $\textbf{d}$ denote its direction. $\textbf{x}$ is fed into a gating network $G$. In Switch-NeRF, $G$ is an MLP network to map point features into gate values $G(\textbf{x})$ for point $\textbf{x}$. In Switch-NeRF++, $G$ is a multi-resolution hash encoding~\cite{mueller2022instant}. It looks up a feature encoding $f_g$ in the hash encoding by computing a hash value of $\textbf{x}$. A small MLP uses $f_g$ to predict gating values $G(\textbf{x})$ for point $\textbf{x}$. According to $G(\textbf{x})$, we use Top-$1$ operation to select an expert $E$ and send $\textbf{x}$ into $E$, which is a larger MLP network in Switch-NeRF and a multi-resolution hash encoding with a distinct resolution range in Switch-NeRF++. $E$ computes a feature $f_e$ for point $\textbf{x}$, which is then passed to a small MLP head to predict the point density $\sigma$ and point color $\textbf{c}$.
We add an additional appearance embedding $\mbox{AE}$~\cite{martin2021nerf, Turki_2022_CVPR} to capture view-dependent appearance when predicting the color.

\subsection{Trainable Gating Network}
\label{hash-based-gating}

The gating network $G$ is pivotal to our method, as it learns the scene decomposition in training. Different previous MoE methods~\cite{DBLP:conf/iclr/ShazeerMMDLHD17, DBLP:conf/iclr/LepikhinLXCFHKS21} that employ different gating networks in different network layers, our network only leverages one gating network. This design choice stems from the fact that applying more gating networks on a massive volume of 3D points in NeRF optimization will largely increase the memory and computation cost. Our experiments also show that more gating networks are not necessary for better decomposition and reconstruction accuracy.

Switch-NeRF employs an MLP network as its gating network.
In contrast, Switch-NeRF++ leverages multi-resolution hash encoding~\cite{mueller2022instant} as $G$, which is a powerful 3D network that can map 3D point $\textbf{x}$ into gate values $G(\textbf{x})$ and learn the scene decomposition more efficiently. Specifically, in Switch-NeRF++, the gating network $G$ consists of a hash operation $\mbox{Hash()}$. For an input 3D point $\textbf{x}$, it first computes a hash index $h(\textbf{x})$ for $\textbf{x}$ and fetches a feature $f_g$ for $\textbf{x}$ from the hash encoding via the index. A small MLP is used to predict gate values $G(\textbf{x})$ from $f_g$. The gate values $G(\textbf{x})$ are normalized into an $n$-dimensional probability distribution via Softmax, with each scalar component $G(\textbf{x})_i$ indicating the likelihood of activating the $i$-th NeRF expert. We derive the hash expert index $i$ corresponding to the chosen expert by applying Top-$1$ operation on this distribution. $\textbf{x}$ is then dispatched into the hash expert $E_i$. 

\noindent The gating procedure of Switch-NeRF is:
\begin{equation}
i, G(\textbf{x})_i = \mbox{Top-1}(\mbox{Softmax}(\mbox{MLP}(\textbf{x}))),
\end{equation}

\noindent The gating procedure of Switch-NeRF++ is:
\begin{equation}
i, G(\textbf{x})_i = \mbox{Top-1}(\mbox{Softmax}(\mbox{MLP}(\mbox{Hash}(\textbf{x})))),
\end{equation}
where $i$ is the selected expert index, and $G(\textbf{x})_i$ is the corresponding gate value. 
The input $\textbf{x}$ is routed to the chosen expert $E_i$, where it is processed to produce the output $E_i(\textbf{x})$. In this way, the gating network can perform a decomposition of the scene into partitions, and each partition can be represented by a distinct expert. 

The expert's final output $\hat{E}(\textbf{x})$ is computed by multiplying the output of $E_i$ with its corresponding gate value: 
\begin{equation}
\vspace{-1pt}
    \hat{E}(\textbf{x}) = G(\textbf{x})_i E_i(\textbf{x}). 
\end{equation}
This multiplication facilitates the gradient flowing through the gating network, enabling joint optimization of both the experts and the gating network during backpropagation. Consequently, the framework can inherently learn scene decomposition through direct gradient-based training.

The use of hash encoding as the gating network in Switch-NeRF++ instead of a pure MLP can provide us with two advantages. The first is that it is much more efficient than a wide MLP head used in previous works~\cite{DBLP:conf/iclr/ShazeerMMDLHD17}. The second is that it is more consistent with our hash experts, and the training of gating aligns well with the hash experts. It also produces clearly higher accuracy.

\begin{figure}[htbp]
\centering
    \includegraphics[width=\linewidth]{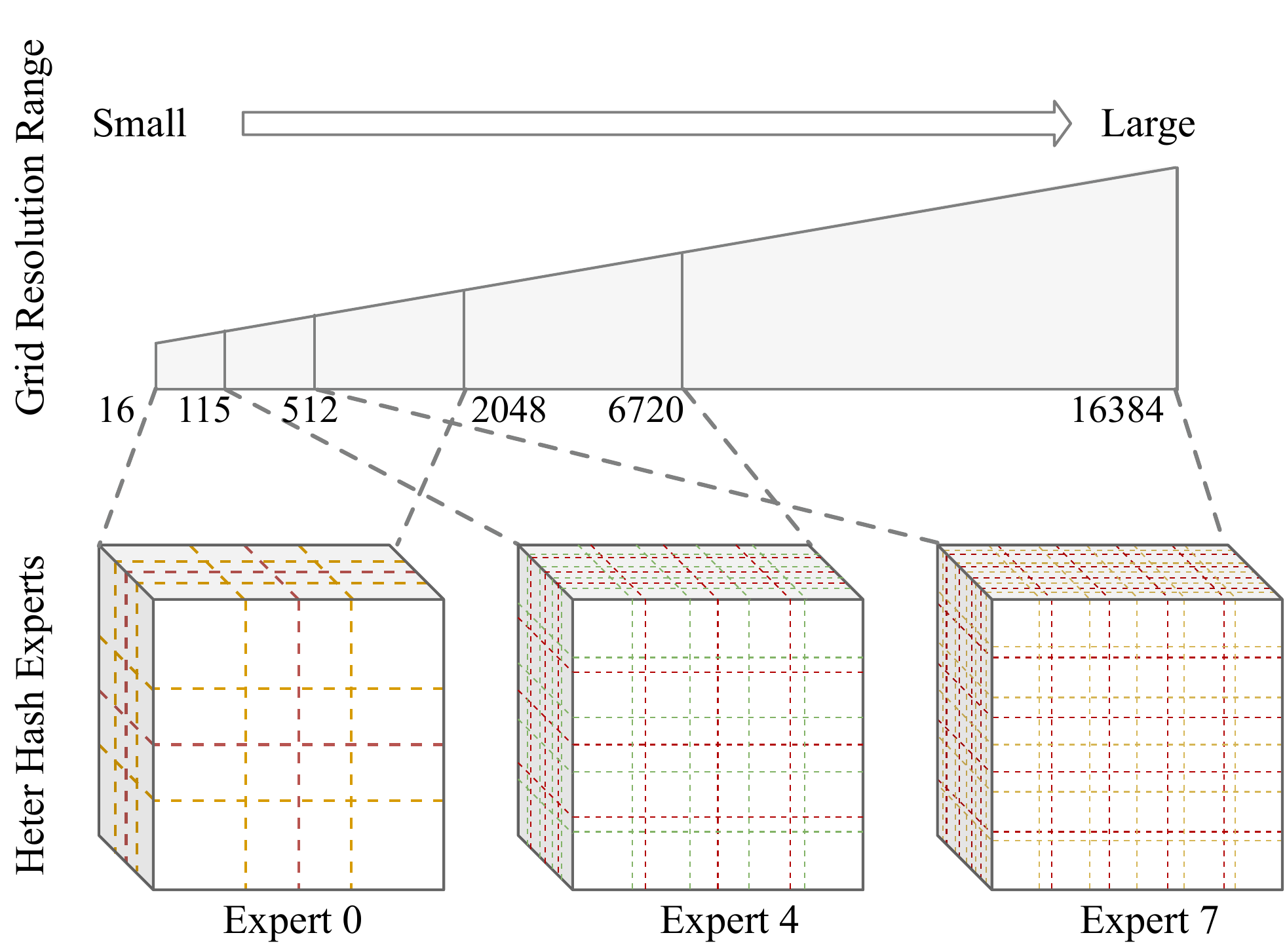}
  \caption{A detailed illustration of our pyramid design of resolution ranges of the heterogeneous experts in Switch-NeRF++. Different experts have different grid resolution ranges from coarse to fine, making the network cover a wide range of resolutions to capture the diverse data distributions. The ranges are controlled by setting the minimum and maximum resolutions for the hash grids in each expert.}
  \label{fig:gate_heter}
\end{figure}

\subsection{Experts for Large Scene Representations}
\label{moHE}

\textbf{MLP-based NeRF Expert Network.} 
Switch-NeRF employs a collection of $n$ MLP experts$\{E_i\}_{i=1}^{n}$, which provide main network parameters to encode scene representation. As illustrated in Fig.~\ref{fig:gateexpert}a, these experts in Switch-NeRF adopt a deep MLP structure with a skip connection, aligning with the core structure design of the vanilla NeRF. The gating network dynamically assigns subsets of 3D points to individual experts. For each 3D point $\textbf{x}$ in an expert $E$, the resulting feature $\hat{E}(\textbf{x})$ is generated by the expert's MLP output multiplied by the corresponding gate value. This feature is then passed to a unified prediction head to drive predictions for both the density $\sigma$ and the color $\textbf{c}$ of $\textbf{x}$.

\noindent \textbf{Hash-based NeRF Expert Network.} In Switch-NeRF++, we efficiently learn the representation of a large-scale scene by our heterogeneous mixture of hash experts (HMoHE). The detailed network structure of the hash expert is shown in Fig.~\ref{fig:gateexpert}b. The hash experts are a set of $n$ hash encodings $\{E_i\}_{i=1}^n$ with different resolution ranges, and they are the main networks to learn a representation of a large-scale 3D scene. For a 3D point $\textbf{x}$, we use the gating network to obtain its corresponding expert $E_i$ and gating value $G_i$. Then, $\textbf{x}$ is input into $E_i$ to extract a feature encoding. The final output of the hash expert $\hat{E}_i(\textbf{x})$ can be written as:
$\hat{E}_i(\textbf{x}) = G_i E_i(\textbf{x})$,
where $G_i$ is used to help train the hash-based gating network. The hash experts are sparse because we only select one expert for a 3D point. This can decouple the network capacity from the computation needed for each point. Since each expert is designed to have different resolution ranges, it can capture the details of regions of quite different distributions. The capacity of the whole network can be easily scaled by adjusting the NeRF expert number $n$. Therefore, our Switch-NeRF++ is highly scalable to model various scales without a large increase in computation. 

\noindent \textbf{Pyramid heterogeneous expert design.} 
In Switch-NeRF, the MLP experts are pure implicit networks that are irrelevant to the actual scene geometry. However, in Swhtch-NeRF++, the hash experts that have multi-resolution grids are highly related to local scene scale. This gives us the possibility to explicitly model the heterogeneity of large-scale scenes by hash experts. Fig.~\ref{fig:gate_heter} shows the design of the heterogeneity of our hash experts. The hash experts are all multi-resolution hash encodings. Instead of using identical experts as other MoE methods~\cite{DBLP:conf/iclr/ShazeerMMDLHD17, mi2023switchnerf}, our different experts have different grid resolution ranges. Each expert has several levels of hash grids. The resolution of each hash grid determines the mapping granularity from a 3D point to a hash index. Experts with coarser hash grids capture more general distributions of the large-scale scene, while those with finer resolutions capture more distinct distributions. To effectively model large-scale scenes, it is crucial to make the network cover a wide range of resolutions to capture the appearance and geometry distribution of the scene accurately. As depicted in Fig.~\ref{fig:gate_heter}, our heterogeneous experts cover a pyramid of resolutions, enabling the gating network to select experts from diverse resolutions, which enhances the representation power and scalability. We control the resolution ranges of experts by setting minimum and maximum resolutions for each expert.

\noindent \textbf{Unified Head.} After generating features of 3D samples from experts, we employ a unified prediction head $H$ to predict the density and color for all the samples. Fig.~\ref{fig:gateexpert}a and Fig.~\ref{fig:gateexpert}b depict the detailed architectures of the unified head in Switch-NeRF and Switch-NeRF++, respectively. The unified head facilitates the sparse framework to encode consistent and unified representations for the different partitions in the entire scene. In the prediction head, a small MLP network is employed to process the expert output $\hat{E}(\textbf{x})$ and predict $\sigma$ and color $\textbf{c}$. The $\hat{E}(\textbf{x})$ is augmented by concatenating a trainable appearance embedding $\mbox{AE}$, to model photometric and environmental variations across training images~\cite{martin2021nerf}.

\subsection{Joint Optimization Losses}

\label{optimizationloss}
We jointly learn the hash-based gating network and HMoHE scene representation with several losses. 
\par\noindent\textbf{Rendering loss.} Our large-scale NeRF framework uses the rendering loss for NeRF optimization~\cite{mildenhall2020nerf} as the primary loss. 
For a training pixel, we sample $N$ points $\{\textbf{x}_i\}_{i=1}^N$ along its viewing ray $\textbf{r}$ and predict the density $\sigma_i$ and color $\textbf{c}_i$ for each point $\textbf{x}_i$ by our network. The final color $\hat{C}(\textbf{r})$ is predicted by the volume rendering~\cite{mildenhall2020nerf} along the ray $\textbf{r}$ as $\hat{C}(\textbf{r}) = \sum_{i=1}^N T_i(1 - \mbox{exp}(-\sigma_i\delta_i))\textbf{c}_i, \text{with}~T_i = \mbox{exp}(-\sum_{j=1}^{i-1}\sigma_j \delta_j)$,
where $\delta_i$ is the distance between neighboring points along $\textbf{r}$. Let $\mathcal{R}$ be the set of rays in a training batch.
The rendering loss $\mathcal{L}_\textbf{r}$ is the Mean Squared Error (MSE) between $\hat{C}(\textbf{r})$ and the ground truth $C(\textbf{r})$:
\begin{equation}
\mathcal{L}_\textbf{r} = \sum_{\textbf{r} \in \mathcal{R}} \left\|\hat{C}(\textbf{r}) - C(\textbf{r})\right\|_2^2.
\end{equation}

\noindent \textbf{Gating balance loss.} Together with $\mathcal{L}_\textbf{r}$, we also employ the gating balance loss commonly used in MoE methods~\cite{DBLP:conf/iclr/LepikhinLXCFHKS21}. This loss encourages the balanced dispatching of points into different hash experts in order to balance the expert training and utilization. Let $n$ be the expert number and $f_i$ be the fraction of points dispatched into an expert $E_i$. We have $\sum{f_i} = 1$. Then, if $\sum{f_i^2}$ is minimized, the gating is ideally balanced. 
In order to make $\sum{f_i^2}$ differentiable for training, as in~\cite{DBLP:conf/iclr/LepikhinLXCFHKS21}, we replace $f_i$ with a soft version $p_i$ which is the fraction of gate values dispatched to $E_i$. The gating balance loss $\mathcal{L}_b$ is defined as
$\mathcal{L}_b = n \sum{f_i g_i}.$
$\mathcal{L}_b$ is minimized as $1$ if the gating is balanced. Let $\lambda$ be the weight for the gating balance loss; the overall loss $\mathcal{L}_o$ is:
\begin{equation}
\mathcal{L}_o = \mathcal{L}_\textbf{r} + \lambda \mathcal{L}_b.
\end{equation}

\subsection{Unbounded Scene}

Real-world large-scale scenes are usually unbounded, which poses challenges in point sampling and grid definition.

For MLP-based Switch-NeRF, we use a method similar to NeRF++~\cite{zhang2020nerf++} and Mega-NeRF~\cite{Turki_2022_CVPR} to divide a large-scale scene into a foreground and a background. In the foreground region, we use cascade sampling similar to NeRF~\cite{mildenhall2020nerf}. In the background region, we use the inverse sphere parameterization similar to NeRF++~\cite{zhang2020nerf++} to handle arbitrary and large depth ranges. 

For hash-based Switch-NeRF++, the hash encodings are defined on a grid in a bounded 3D space. In the foreground region, we use the normal hash grids. The background uses a contraction method in Mip-NeRF 360~\cite{Barron_2022_CVPR} to map the unbounded space into a bounded space.
In the contraction, we normalize the scene into a unit sphere with the scene bound, and then apply the contraction on a 3D point $\textbf{x}$. If $\|\textbf{x}\| \leq 1$, we do not change its value. If $\|\textbf{x}\| > 1$, we transform $\|\textbf{x}\|$ by $\left(2-\frac{1}{\|\textbf{x}\|}\right)\left(\frac{\textbf{x}}{\|\textbf{x}\|}\right)$.
This contraction function maps the background 3D points into a bounded scene without affecting the
3D points in the foreground. The hash grids for the background are defined in the contracted space.

\subsection{Point dispatching}

In the gating operation, after we obtain the expert indexes for the input points, we need to dispatch them to different experts. The MLP-based gating and the hash-based gating are different from each other in the network operation. Therefore, we have different dispatching designs for different gating network to maximize their efficiency.

\subsubsection{Point dispatching in MLP-based gating.} 
\label{sec:fulldispatch}

\textbf{Capacity factor for training.} In MLP-based gating, it is important to efficiently reorganize and send a large number of 3D points to different experts. The dispatching method using Einops-based operations~\cite{DBLP:conf/iclr/LepikhinLXCFHKS21} usually struggles with memory overflow due to the massive volume of 3D points. In the training of Switch-NeRF, we adopt the optimized CUDA dispatch operation~\cite{hwang2022tutel} for training efficiency. We introduce a capacity factor $C_f$ that limits the input sample number per NeRF expert in the training, following the setting of previous MoE methods~\cite{DBLP:conf/iclr/LepikhinLXCFHKS21}. This ensures uniform tensor dimensions, balanced computation, and predictable communication overheads. Given a total batch size $B$, $n$ NeRF experts, the maximum number of sample points per NeRF expert is $B_e = \mbox{ceil}(\frac{kBC_f}{n})$, where $k$ is the number of chosen experts for each sample.

\begin{figure}[htbp]
\centering
    \includegraphics[width=\linewidth]{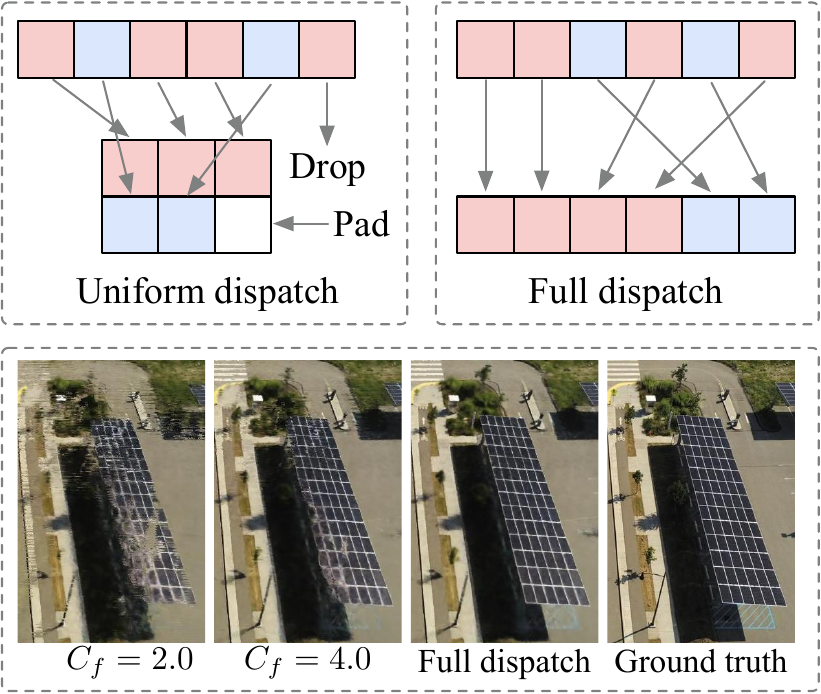}
  \caption{The comparison between the uniform dispatch and full dispatch in Switch-NeRF. In training, the uniform dispatch is utilized for better computation and communication efficiency. The input tensor of each expert has the same shape. To maintain the shape, it will drop overflow tokens or pad tokens if necessary. In inference, the full dispatch is employed to make sure each point will be processed by an expert, avoiding accuracy decrease. If the uniform dispatch is used in inference, the rendered images will have artifacts.}
  \label{fig:dispatchcompare}
\end{figure}

This uniform dispatch with a capacity factor is depicted in Fig. \ref{fig:dispatchcompare}. The figure shows that, when $C_f = 1.0$, overflow points for an expert are dropped, and if the expert capacity is not fully used, it will be zero-padded, ensuring uniform tensor shapes.
Increasing $C_f$ will decrease the dropping ratio but raise memory and computation cost. In the training of Switch-NeRF, we empirically set $C_f$ to 1.0 to avoid consuming extra memory. To further mitigate performance loss from lower capacities, we integrate the Batch Prioritized Routing~\cite{DBLP:conf/nips/RiquelmePMNJPKH21} method, which refines training by prioritizing critical samples.

\noindent \textbf{Full dispatch in inference.} While existing MoE methods~\cite{DBLP:conf/iclr/LepikhinLXCFHKS21, fedus2022switch} typically employ the uniform dispatch for both training and testing, this uniform approach may inevitably drop sample points in NeRF inference. In Switch-NeRF's MLP-based architecture, we find that the uniform dispatch works well in the training but introduces notable accuracy degradation during inference due to dropped sample points. We attribute this to our design of using only one gate network instead of stacked MoE layers and skip connections to maintain the network efficiency. This may make our MoE structure less robust to point dropping in inference.
Our solution addressing this problem in inference is to introduce an optimized full dispatch strategy leveraging Pytorch, CUDA, and Tutel. This efficient strategy dispatches all the input points to their corresponding expert with marginal memory overhead, as illustrated in Fig.~\ref{fig:dispatchcompare}. This strategy yields significant accuracy improvements in final model inference. The full dispatch operation is also explored in recent large language models (LLMs) based on MoE. Deepseek-V3~\cite{liu2024deepseek} describes a similar strategy, where the model maintains good load balance and avoids token dropping throughout both training and inference. This approach aligns closely with our motivation for using full dispatch at inference time, especially in the context of 3D rendering.

\subsubsection{Efficient hash dispatching in hash gating network.} 

The MLP gating network used in Switch-NeRF needs extra memory and computation for dispatching because it performs matrix multiplication on features. Different from this, the hash-based gating network fetches features from the hash tables without matrix operations. It first computes a hash index for each 3D point. Then it computes hash table offsets by the hash index and fetches features from the hash tables. In our implementation, we find it is very straightforward to integrate the dispatching into this hash operation. We directly pass the expert indexes into the hash function and use them to compute the hash table offset, which adds only minimal computation and does not require extra memory to reorganize the inputs. We do not need any explicit rearrangement of the input features and do not need to consider the capacity factor. Our implementation of hash-based dispatching can further boost the efficiency of our Switch-NeRF++ framework.

\subsection{Method conclusion.} 

\begin{figure}[htbp]
\centering
    \includegraphics[width=\linewidth]{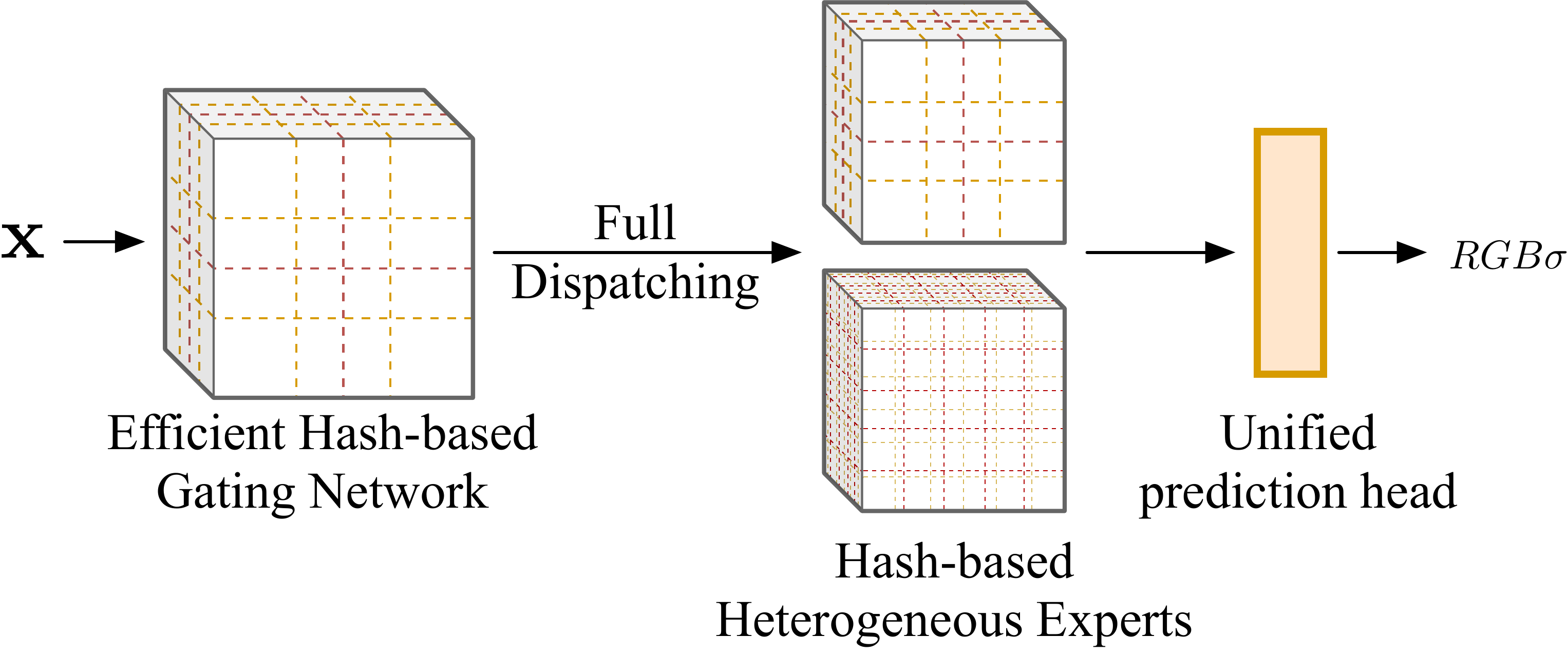}
  \caption{Block diagram illustrating the integrated architectural innovations in Switch-NeRF++. With these innovations, our method exhibits significant efficiency gains, with an 8x acceleration in training and a 16x acceleration in rendering compared to Switch-NeRF.}
  \label{fig:reviewer1_blocks}
\end{figure}

In conclusion, Switch-NeRF++ integrates a set of architectural innovations that collectively improve both accuracy and efficiency over the original Switch-NeRF. As shown in Fig.~\ref{fig:reviewer1_blocks}, the efficient hash gating network combined with full dispatching ensures both the speed and accuracy of expert selection. The heterogeneous hash experts provide accurate and compact scene representations. The innovations in gating, dispatching, and experts play a critical role in speeding up the training and inference. The unified prediction head (MLP) ensures consistency across different experts. Together with these innovations, the Switch-NeRF is upgraded into Switch-NeRF++ with substantial improvements in efficiency and accuracy.

\section{Experiments}
\subsection{Datasets and Metrics}

\textbf{Datasets.} To thoroughly assess the scalability, accuracy, and efficiency of our method, we conduct experiments on four large-scale datasets, \emph{i.e.}, Mega-NeRF \cite{Turki_2022_CVPR}, Block-NeRF~\cite{Tancik_2022_CVPR}, an adapted version of UrbanBIS~\cite{UrbanBISbib}, and the MatrixCity~\cite{li2023matrixcity}. The Mega-NeRF dataset comprises 5 scenes (Building, Rubble, Residence, Sci-Art, and Campus), each with $2k$ to $6k$ images of $4k$ to $5k$ resolution, covering areas from 0.15 to 1.3 $km^2$. The Block-NeRF dataset includes a scene with $12k$ images at $1k \times 1k$ resolution, covering a $1km$ Urban street. To fully validate the scalability of our method, we adapt a much larger scene (Longhua) from UrbanBIS~\cite{UrbanBISbib} with $13k$ images at $8k \times 5k$ resolution, covering a $6.5km^2$ Urban area. We estimate camera parameters by colmap~\cite{schoenberger2016sfm} and downsample the UrbanBIS images by 4. 
The MatrixCity~\cite{li2023matrixcity} aerial dataset contains 5 blocks, each with thousands of high-resolution synthetic images.
These scenes are very challenging due to the huge image data and the coverage of large-scale regions, suitable for evaluating the scalability of our method.

\noindent \textbf{Metrics.} The standard metrics such as PSNR, SSIM~\cite{wang2004image} (both higher is better) and LPIPS~\cite{zhang2018unreasonable} (lower is better) the same as Mega-NeRF~\cite{Turki_2022_CVPR} are used to evaluate the rendering accuracy.

\noindent \textbf{Visualization.} In addition to visualizing the rendered 2D images, we also employ a 3D visualization strategy to directly visualize the 3D radiance fields. 3D points are sampled along pixel rays and merged into a 3D point cloud. Each 3D point uses the $\alpha = 1-\mbox{exp}(-\sigma_i \delta_i)$ as its opacity. We leverage the Point Cloud Library~\cite{rusu20113d} to render both the color and opacity of these 3D points to visualize the radiance fields.

\subsection{Implementation Details}

\begin{algorithm}[t]
\caption{The forward and loss computation of Switch-NeRF++.}
\label{alg:switchnerfpp}
\begin{algorithmic}[1]
{\REQUIRE \textbf{Inputs:} 3D points $\mathbf{x}$, view directions $\mathbf{d}$, appearance embeddings $\mathbf{a}$

\STATE \textbf{// === Part I: Network Configuration ===}
\STATE \textbf{// --- Hash Gating Network ---}
\STATE Define \texttt{base\_resolution}: $16$
\STATE Define \texttt{desired\_resolution}: $2048$
\STATE Define \texttt{hashmap\_size}: $2^{19}$
\STATE $\text{HashGate} = \text{HashEncoding}(16^3, 2048^3)$
\STATE \textbf{// --- Hash Experts ---}
\STATE Define number of hash experts $N=8$
\STATE Define a set of \texttt{base\_resolution} $\{r\}$, logarithmically spaced in $[16, 512]$ for $N=8$ hash experts
\STATE Define a set of \texttt{desired\_resolution} $\{\tilde{r}\} $, logarithmically spaced in $[2048, 16384]$ for $N=8$ hash experts
\FOR{$i = 1$ to $N$}
    \STATE $\text{HashExpert}_i = \text{HashEncoding}({\{r\}_i}^3, {\{\tilde{r}\}_i}^3)$
\ENDFOR
\STATE The resolution intervals are logarithmically sampled to cover the scale distribution of urban scenes.

\STATE \textbf{// === Part II: Forward Pass ===} 
\STATE \textbf{\textit{// --- Gating Network ---}}
\STATE Encode position: $\mathbf{f}_g \leftarrow \text{HashGate}(\mathbf{x})$
\STATE Compute gate values: $\mathbf{G} = \text{Softmax}(\mathbf{f}_g)$
\STATE Select top-1 hash expert: $i = \arg\max(\mathbf{G})$

\STATE \textbf{\textit{// --- Hash Expert Processing ---}}
\STATE Hash Expert encoding: $\mathbf{E}_i(\mathbf{x}) \leftarrow \text{HashExpert}_i(\mathbf{x})$
\STATE Apply gate value: $\hat{\mathbf{E}}_i(\mathbf{x}) = \mathbf{G}_i \cdot \mathbf{E}_i(\mathbf{x})$
\STATE \textbf{\textit{// --- Prediction ---}}
\STATE Direction encoding: $\mathbf{SH}_d = \text{SphericalHarmonics}(\mathbf{d})$
\STATE Concatenate: $\mathbf{f}_{\text{final}} = [\hat{\mathbf{E}}_i(\mathbf{x}), \mathbf{SH}_d, \mathbf{a}]$
\STATE Predict density: $\sigma = \text{MLP}_\text{density}(\hat{\mathbf{E}}_i(\mathbf{x}))$

\STATE Predict color: $\mathbf{c} = \text{MLP}_\text{color}(\hat{\mathbf{E}}_i(\mathbf{x}))$

\STATE \textbf{\textit{// Rendering}}

\STATE $T_i = \mbox{exp}(-\sum_{j=1}^{i-1}\sigma_j \delta_j)$
\STATE Rendered Color: $\hat{C}(\textbf{r}) = \sum_{i=1}^N T_i(1 - \mbox{exp}(-\sigma_i\delta_i))\textbf{c}_i$

\STATE \textbf{\textit{// Losses}}
\STATE Ground Truth Color: $C(\textbf{r})$
\STATE Rendering Loss: $\mathcal{L}_\textbf{r} = \sum_{\textbf{r} \in \mathcal{R}} \left\|\hat{C}(\textbf{r}) - C(\textbf{r})\right\|_2^2$

\STATE Gate Balance Loss: $\mathcal{L}_b = n \sum{f_i g_i}.$
\STATE Balance Loss Weight: $\lambda = 5 \times 10^{-4}$
\STATE Overall loss: $\mathcal{L}_o = \mathcal{L}_\textbf{r} + \lambda \mathcal{L}_b$

}

\end{algorithmic}
\end{algorithm}

In the MLP-based Switch-NeRF, the foreground network contains 8 NeRF experts configured with a Top-$1$ gating network and a capacity factor $C_f=1.0$, while the background is represented by a single NeRF network. Each NeRF expert is structured as a 7-layer MLP with 256 channels per layer. In training and inference, 256 coarse and 512 ﬁne points are sampled per ray for the foreground network, whereas 128 coarse and 256 fine samples are sampled for the background network. The training utilizes 8 NVIDIA RTX 3090 GPUs, with a global batch size of 8192 rays in each iteration. The models are optimized by 500k iterations for each large-scale scene and then evaluated on validation images.

In our Switch-NeRF++ framework, we utilize multi-resolution hash encoding with 16 levels, each level with 2 feature channels. The maximum hash map size is $2^{19}$. For the Mega-NeRF and Matrixcity datasets, we use the same hash experts. For the larger Block-NeRF dataset and UrbanBIS dataset, we validate our design of a heterogeneous mixture of hash experts (HMoHE) using both the same hash experts and heterogeneous hash experts. The hash gating network includes a multi-resolution hash encoding and a 3-layer MLP with 64 channels. We set 8 hash experts in our experiments for the foreground and one hash encoding for the background. The $\sigma$ prediction head is a 2-layer MLP and the color prediction head is a 3-layer MLP, both with 64 channels. We conduct our experiments on 4 NVIDIA RTX 3090 GPUs, with a batch size of 16384 rays per GPU. We use larger batch sizes to fully use the computation power of the GPUs since our network uses much less memory than MLP-based NeRF methods. We use coarse-to-fine sampling similar to Mega-NeRF, sampling 128 coarse samples and 128 fine samples along a ray in the foreground and half of them in the background. 

We use Adam optimizer to train Switch-NeRF and Switch-NeRF++ with a learning rate of $5 \times 10^{-4}$ that exponentially decreases. The weight $\lambda$ of balance loss $\mathcal{L}_b$ is set as $5 \times 10^{-4}$ for all experiments. We also use half-precision training and testing, similar to Mega-NeRF.

The detailed configuration and training pipeline of Switch-NeRF++ are shown in Algorithm~\ref{alg:switchnerfpp}. We configure the network with a hash-based gating module following the default Instant-NGP settings, and define $N=8$ hash experts with logarithmically spaced base and desired resolutions, enabling each expert to specialize in different scene scales. For each 3D point $\mathbf{x}$, it is processed by the gating network, one of the hash experts, and prediction heads to predict density and color. Volume rendering is applied along the ray to render the pixel color and compute $L_2$ loss. A gating balance loss is also used to balance the training.

\noindent \textbf{Baseline.} We compare our method with the large-scale NeRF method Mega-NeRF~\cite{Turki_2022_CVPR}. We also adapt the Instant-NGP~\cite{mueller2022instant} into a large-scale setting, named INGP. We use a foreground and a background network in the baseline. Each of them consists of one multi-resolution hash encoding and prediction heads. We set the hash map size as $2^{22}$ in the foreground network so it has the \textbf{same} network capacity as our method. The sampling strategy is the same as our Switch-NeRF++.

\begin{figure*}[htbp]
\centering
    \includegraphics[width=\textwidth]{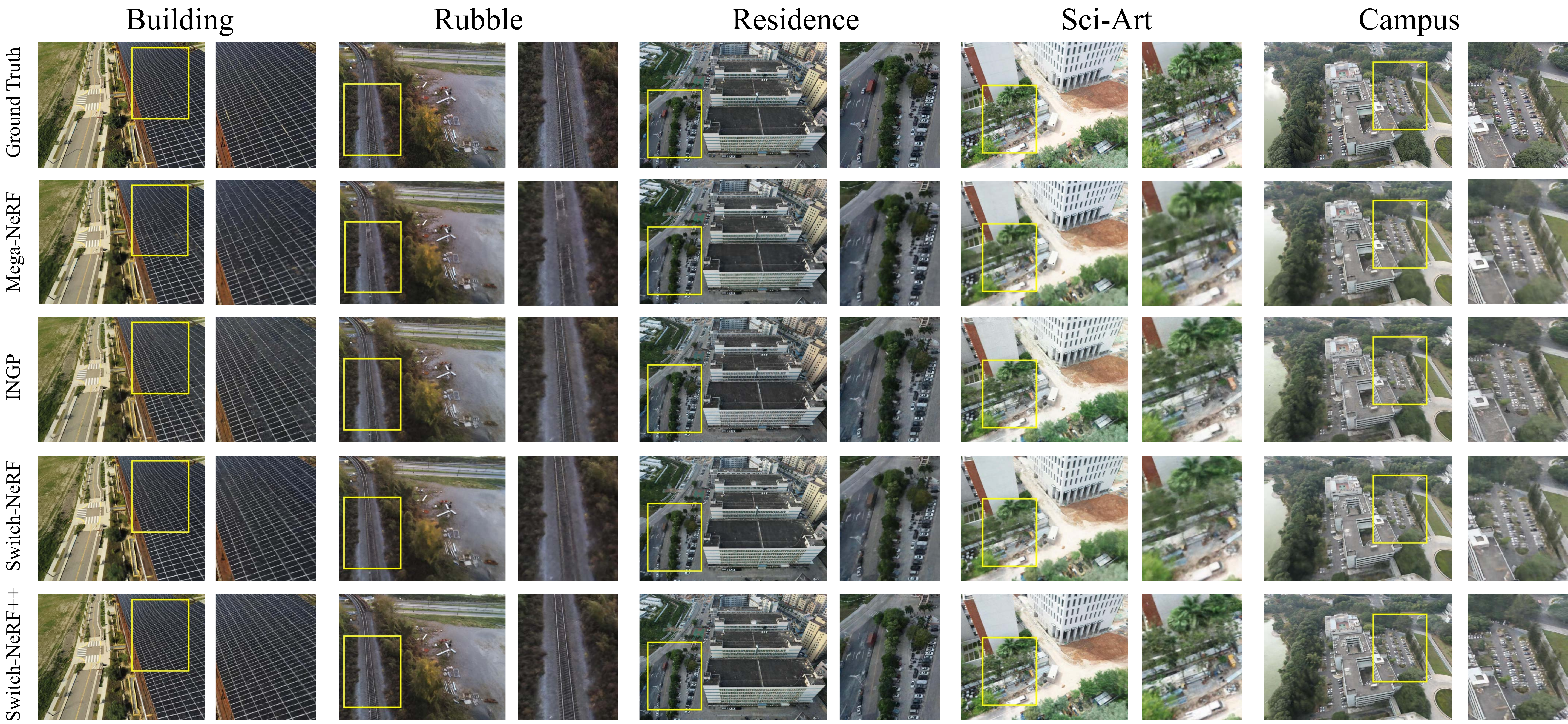}
  \caption{A qualitative comparison of rendered images of the INGP~\cite{mueller2022instant}, Mega-NeRF~\cite{Turki_2022_CVPR}, Switch-NeRF~\cite{mi2023switchnerf}, and our Switch-NeRF++ on Mega-NeRF dataset. Our Switch-NeRF++ framework can render cleaner and sharper details compared to the state-of-the-art methods.}

  \label{fig:imagecompare}
\end{figure*}

\begin{table*}[htbp]
  \centering
  \caption{The benchmark performances of our Switch-NeRF++ on the Mega-NeRF~\cite{Turki_2022_CVPR} dataset. In this table, the INGP is the Instant-NGP~\cite{mueller2022instant} trained with 100k steps. Switch-NeRF++* is trained with 50k steps, and Switch-NeRF++ is trained with 100k steps. Our method can achieve state-of-the-art results with only about a training time of 5.5h, which is an \textbf{8x} acceleration compared to Switch-NeRF~\cite{mi2023switchnerf} and a \textbf{5x} acceleration compared to Mega-NeRF~\cite{Turki_2022_CVPR}.}
    \resizebox{\linewidth}{!}{
    \begin{tabular}{c|ccc|ccc|ccc|ccc|ccc|c}
    \toprule
    Dataset & \multicolumn{3}{c|}{Building} & \multicolumn{3}{c|}{Rubble} & \multicolumn{3}{c|}{Residence} & \multicolumn{3}{c|}{Sci-Art} & \multicolumn{3}{c|}{Campus} &  \\
    \midrule
    Metrics & PSNR  & SSIM  & LPIPS & PSNR  & SSIM  & LPIPS & PSNR  & SSIM  & LPIPS & PSNR  & SSIM  & LPIPS & PSNR  & SSIM  & LPIPS & Time \\
    \midrule
    INGP  & 20.59 & 0.555 & 0.480 & 24.95 & 0.627 & 0.443 & 22.22 & 0.679 & 0.422 & 24.52 & 0.778 & 0.379 & 23.50 & 0.567 & 0.568 & 10.3h \\
    Mega-NeRF  & 20.93 & 0.547 & 0.504 & 24.06 & 0.553 & 0.516 & 22.08 & 0.628 & 0.489 & 25.60 & 0.770 & 0.390 & 23.42 & 0.537 & 0.618 & 31.4h \\
    Switch-NeRF & 21.54 & 0.579 & 0.474 & 24.31 & 0.562 & 0.496 & 22.57 & 0.654 & 0.457 & \textbf{26.52} & 0.795 & 0.360 & 23.62 & 0.541 & 0.609 & 43.0h \\
    Switch-NeRF++* & 21.27 & 0.595 & 0.447 & 25.11 & 0.642 & 0.424 & 22.79 & 0.704 & 0.403 & 25.65 & 0.807 & 0.348 & 23.54 & 0.565 & 0.570 & 5.50h \\
    Switch-NeRF++ & \textbf{21.79} & \textbf{0.622} & \textbf{0.424} & \textbf{25.55} & \textbf{0.668} & \textbf{0.400} & \textbf{23.25} & \textbf{0.725} & \textbf{0.382} & 26.17 & \textbf{0.821} & \textbf{0.329} & \textbf{23.94} & \textbf{0.584} & \textbf{0.545} & 11.0h \\
    \bottomrule
    \end{tabular}%
    }
    
  \label{table:magadataresults}%
\end{table*}%

\subsection{Overall Benchmark Results}
\noindent \textbf{Mega-NeRF dataset.} We evaluate the accuracy and efficiency of our framework on Mega-NeRF~\cite{Turki_2022_CVPR} dataset.
The accuracy and training time are reported in Table~\ref{table:magadataresults}. The results of Mega-NeRF is quoted from their papers. 
In Table~\ref{table:magadataresults}, Mega-NeRF and Switch-NeRF are trained with about 30h and 44h, respectively, for 500k steps. Instant-NGP (INGP)~\cite{mueller2022instant} is trained with about 10h for 100k steps. Switch-NeRF++* is our method trained with 5.5h for 50k steps. Switch-NeRF++ is our method trained with 11h for 100k steps. The results show that our MLP-based Switch-NeRF can clearly outperform its counterpart Mega-NeRF, validating the effectiveness of our learnable decomposition and Mixture of Experts. Switch-NeRF++* trained with only 5.5 hours can already achieve state-of-the-art results on almost all the metrics. This is an acceleration of $8\times$ compared to Switch-NeRF and $5\times$ compared to Mega-NeRF, proving the superior efficiency and accuracy of our method. Our Switch-NeRF++ clearly outperforms INGP with less training time, showing the great scalability of our Switch-NeRF++ framework. The training time of our method is comparable to INGP with the same steps. One reason is that we implement an efficient dispatching method by integrating point dispatching into the hash operation. Another reason is that we set a similar number of parameters in INGP as in our network. We provide qualitative results in Fig.~\ref{fig:imagecompare}. Our method can render images of higher quality with cleaner and sharper details. 
The main results on the Mega-NeRF dataset clearly demonstrate that our Switch-NeRF++ can achieve better accuracy with superior efficiency compared to Mega-NeRF, Switch-NeRF, and INGP.

\begin{figure*}[htbp]

\centering
    \includegraphics[width=\linewidth]{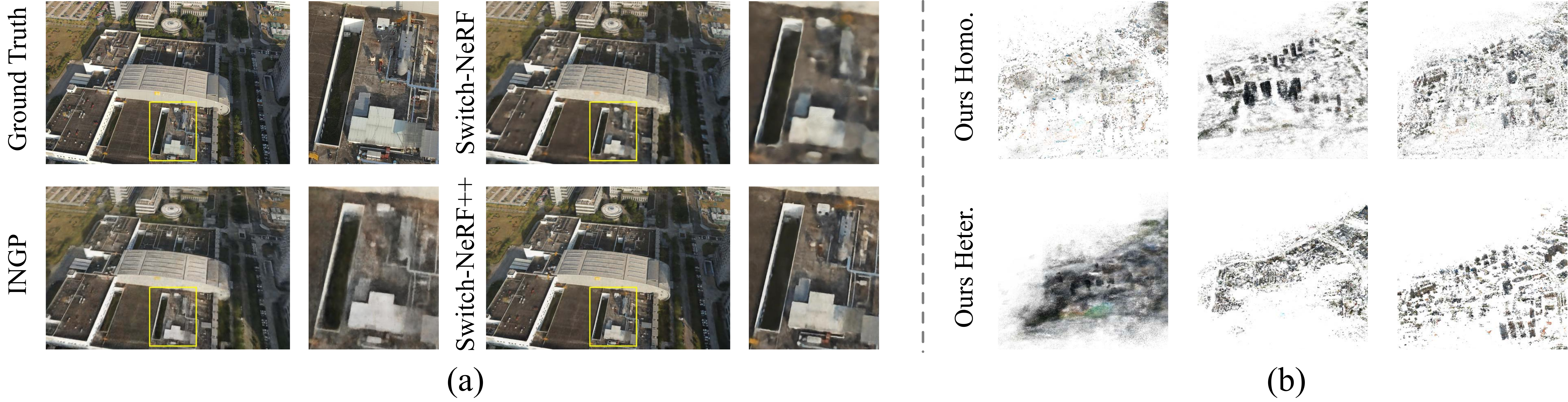}

  \caption{(a) Rendered images of INGP~\cite{mueller2022instant}, Switch-NeRF~\cite{mi2023switchnerf}, and our Switch-NeRF++ on UrbanBIS. Ours can render images with notably higher quality compared to other methods. (b) A visualization of our decomposition of the radiance field of the UrbanBIS dataset. Different experts of Ours (Heter.) capture quite different distributions of the scene, from coarser grounds to finer buildings. Different experts in Ours (Homo.) handle similar distributions. This figure shows that Ours (Heter.) can learn to capture heterogeneous representations effectively.}
    \vspace{-15pt}
  \label{fig:image_UrbanBIS_point}
\end{figure*}

\noindent \textbf{UrbanBIS dataset.} We evaluate the accuracy and scalability of our method on the very large UrbanBIS~\cite{UrbanBISbib} dataset. Table~\ref{table:BlockUrbanBISdataresults} demonstrates that our heterogeneous hash experts (Heter.) significantly outperform all the compared methods. Notably, our heterogeneous version shows substantial improvements over our homogeneous version (Homo.) across all metrics (PSNR, SSIM, and LPIPS), emphasizing the importance of considering scene heterogeneity when scaling to larger scenes. These experiments strongly support the scalability of our design of HMoHE in Switch-NeRF++ when applied to very large-scale scenes. Fig.~\ref{fig:image_UrbanBIS_point}(a) illustrates that our method produces images with enhanced sharpness and details.

\begin{table}[htbp]
  \centering
\caption{The benchmark performance of Switch-NeRF++ on UrbanBIS~\cite{UrbanBISbib} and Block-NeRF~\cite{Tancik_2022_CVPR} dataset. Switch-NeRF++ (Heter.) achieves significant improvements over all compared methods across all metrics, including our homogeneous version (Homo.). The results highlight the good scalability of our method and the importance of heterogeneous modeling for achieving scalability.
}
\vspace{3pt}
\resizebox{\linewidth}{!}{

    \begin{tabular}{c|c|cccc}
    \toprule[1pt]
    Dataset & Metrics & Switch-NeRF & INGP &  Switch-NeRF++ (Homo.) & Switch-NeRF++ (Heter.) \\
    \midrule
    \multirow{4}[0]{*}{UrbanBIS} 
    & PSNR$\uparrow$ & 19.49 & 19.58 & 20.42 & \textbf{20.76} \\
    & SSIM$\uparrow$ & 0.412 & 0.427 & 0.458 & \textbf{0.480} \\
    & LPIPS$\downarrow$ & 0.714 & 0.695 & 0.655 & \textbf{0.628} \\
    \midrule
    \multirow{4}[0]{*}{Block-NeRF } 
    & PSNR$\uparrow$ & 23.86 & 22.10 & 23.90 & \textbf{24.11} \\
    & SSIM$\uparrow$ & 0.762 & 0.757 & 0.790 & \textbf{0.795}  \\
    & LPIPS$\downarrow$ & 0.489 & 0.490 & 0.454 & \textbf{0.448} \\
    \bottomrule
    \end{tabular}
}
\label{table:BlockUrbanBISdataresults}
\vspace{-5pt}
\end{table}%

\noindent \textbf{Block-NeRF dataset.} We evaluate the accuracy and scalability of our method on the large-scale Block-NeRF dataset~\cite{Tancik_2022_CVPR}. Quantitative results are in Table~\ref{table:BlockUrbanBISdataresults}. Our Switch-NeRF++, employing heterogeneous hash experts (Heter.), outperforms INGP~\cite{mueller2022instant} and Switch-NeRF~\cite{mi2023switchnerf} by considerable margins, particularly in terms of SSIM and LPIPS. This highlights its potential for high-fidelity scene reconstruction. Moreover, our heterogeneous design of hash experts consistently outperforms the homogeneous version (Homo.), showing the importance of heterogeneous modeling of very large scenes.

\noindent \textbf{MatrixCity dataset.}
Our method on MatrixCity~\cite{li2023matrixcity} in Table~\ref{tab:accuracymatrixcity} achieves an overall \textbf{1.75-point} improvement in PSNR. The results of NeRF~\cite{mildenhall2020nerf}, Instant-NGP (INGP)~\cite{mueller2022instant} and Mip-360~\cite{Barron_2022_CVPR} are quoted from the MatrixCity. Note that the parameters of INGP in MatrixCity aligned with our network, using $2^{22}$ entries as indicated in MatrixCity's supplementary file. These experiments demonstrate superior accuracy and general applicability of our method across different types of scenes.

\subsection{Model Analysis}

\begin{figure}[htbp]

\centering
    \includegraphics[width=\linewidth]{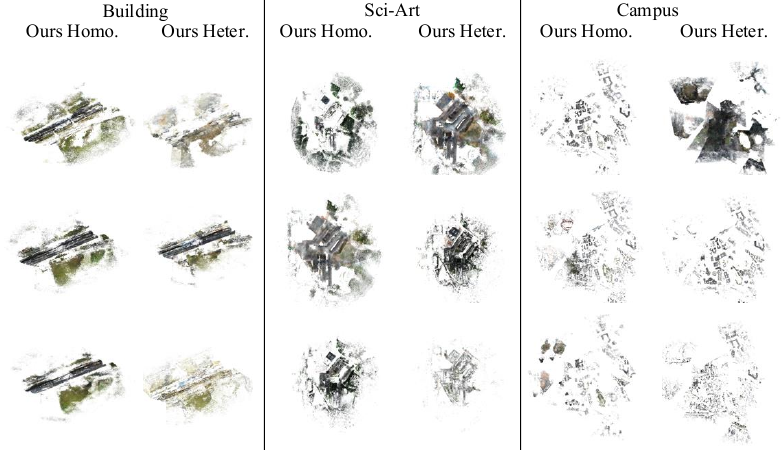}

  \caption{Visualization results of our decomposition of the radiance field on the Mega-NeRF datasets. Our heterogeneous designs can capture quite different distributions across different scenes, from coarser to finer structures, while our homogeneous designs handle similar distributions.}
  \label{fig:expert_decomposition}
\end{figure}

\noindent \textbf{Effect of heterogeneous mixture of hash experts.} We validate our heterogeneous hash design on Block-NeRF~\cite{Tancik_2022_CVPR} and UrbanBIS~\cite{UrbanBISbib} datasets. The scene in the Block-NeRF dataset covers a $1km$ street. The scene in the UrbanBIS dataset covers a $6.5km^2$ region. Both scenes are large enough to evaluate the scalability of our heterogeneous design. The results in Table~\ref{table:BlockUrbanBISdataresults} clearly demonstrate the superior performance of our heterogeneous design compared to our homogeneous version (Homo.), particularly on UrbanBIS. Each of the experts in Switch-NeRF++ (Homo.) uses the default setting of the encodings in Instant-NGP~\cite{mueller2022instant}. In contrast, Switch-NeRF++ (Heter.) incorporates experts with varying resolution ranges, with minimum resolutions of the experts exponentially increasing from 16 to 512 and maximum resolutions exponentially increasing from 2048 to 16384. The ablation results strongly emphasize the significance of considering scene heterogeneity when scaling to larger scenes and provide clear evidence for the scalability of our method. Fig.~\ref{fig:image_UrbanBIS_point}(b) visualizes the decomposition of the radiance field of UrbanBIS dataset of Switch-NeRF++ (Homo.). and Switch-NeRF++ (Heter.), and Fig.~\ref{fig:expert_decomposition} visualizes the decomposition of the Mega-NeRF dataset. As shown in these figures, the point clouds of Switch-NeRF++ (Homo.) have similar or random sharpness, showing that their decomposition is not related to local scales. In contrast, the point clouds of Switch-NeRF++ (Heter.) demonstrate a pattern of increasing sharpness along with the increase of resolution. These visualizations further validate that our heterogeneous design can effectively capture the different distributions of the large-scale scene.

\begin{table}[htbp]
\centering
\caption{PNSR on MatrixCity~\cite{li2023matrixcity} dataset. 
Switch-NeRF++ achieves an overall \textbf{1.75-point} improvement in PSNR. The results of other methods are quoted from the MatrixCity paper. Please note that INGP also uses $2^{22}$ entries, whose number of parameters aligns with our network. These experiments demonstrate that our method can handle different types of scenes.}
\vspace{4pt}
\resizebox{\linewidth}{!}{

    \begin{tabular}{c|ccccc|c}
    \toprule
          & Block A & Block B & Block C & Block D & Block E & Overall \\
    \midrule
    NeRF  & 23.15 & 22.94 & 22.15 & 23.09 & 23.53 & 22.97 \\
    INGP  & 27.21 & 25.45 & 23.21 & 26.24 & 26.36 & 25.69 \\
    Mip-360 & 26.64 & 24.80  & 24.20  & 26.45 & 26.54 & 25.73 \\
    \midrule
    \multirow{1.5}[2]{*}{Switch-NeRF++}  
    &   \textbf{28.15} &  \textbf{27.47}     &   \textbf{26.72}   & \textbf{27.56} &   \textbf{27.50}    & \textbf{27.48} \\
    &   \textbf{{+0.94}}  &   \textbf{{+2.01}}    &  \textbf{{+2.52}}   & \textbf{{+1.11}} &   \textbf{{+0.96}}    &  \textbf{{+1.75}} \\
    \bottomrule
    \end{tabular}%
    }
\label{tab:accuracymatrixcity}
\end{table}

\begin{table}[htbp]
\centering
\caption{Ablation studies of our key designs in Switch-NeRF++. The MLP-gating does not perform as well as our proposed hash-based gating network. With 2 experts, our method produces good accuracy, but it is lower than the accuracy of the one with 8 experts. The Top-2 operation can obtain comparable results to the Top-1 operation.
}

\resizebox{\linewidth}{!}{

    \begin{tabular}{c|ccc|ccc}
    \toprule
    Dataset & \multicolumn{3}{c|}{Residence} & \multicolumn{3}{c}{Sci-Art} \\
    \midrule
    Metrics & PSNR$\uparrow$ & SSIM$\uparrow$ & LPIPS$\downarrow$ & PSNR$\uparrow$ & SSIM$\uparrow$ & LPIPS$\downarrow$ \\
    \midrule
    MLP-Gate & 22.03 & 0.658 & 0.443 & 24.00 & 0.761 & 0.399 \\
    2 Experts & 22.62 & 0.683 & 0.429 & 25.61 & 0.799 & 0.362 \\
    Top-2 & 23.33 & 0.731 & 0.374 & 26.08 & 0.819 & 0.328 \\
    Top-1 & 23.25 & 0.725 & 0.382 & 26.17 & 0.821 & 0.329 \\
    \midrule
    \end{tabular}%
    }
\label{table:ablationstudy}
\end{table}

\begin{table}[htbp]
\centering
\caption{Comparison to GP-NeRF~\cite{zhang2023efficient}. Switch-NeRF++ consistently outperforms GP-NeRF, especially on the very large UrbanBIS~\cite{UrbanBISbib} dataset. GP-NeRF~\cite{zhang2023efficient} uses hash encodings together with 2D ground plane features without scene decomposition. Our learnable heterogeneous decomposition on large-scale scenes better exploits network parameters.
}
\vspace{2pt}
\resizebox{\linewidth}{!}{

    \begin{tabular}{c|c|cc||c|c|cc}
    \toprule[1pt]
    Dataset & Metrics & GP-NeRF & Switch-NeRF++ & Dataset & Metrics & GP-NeRF & Switch-NeRF++\\
    \midrule
    \multirow{3}[0]{*}{UrbanBIS} 
          & PSNR$\uparrow$ & 19.51 & \textbf{20.76} & \multirow{3}[0]{*}{Rubble } 
          & PSNR$\uparrow$ & 24.84 & \textbf{25.55} \\
          & SSIM$\uparrow$ & 0.426 & \textbf{0.480} & & SSIM$\uparrow$ & 0.611 & \textbf{0.668} \\
          & LPIPS$\downarrow$ & 0.695 & \textbf{0.628} & & LPIPS$\downarrow$  & 0.448 & \textbf{0.400}\\
    \midrule
    \multirow{3}[0]{*}{Building} 
          & PSNR$\uparrow$ & 21.42 & \textbf{21.79} & \multirow{3}[0]{*}{Residence} 
          & PSNR$\uparrow$ & 22.84 & \textbf{23.25} \\
          & SSIM$\uparrow$ & 0.602 & \textbf{0.622} & & SSIM$\uparrow$ & 0.703 & \textbf{0.725}\\
          & LPIPS$\downarrow$ & 0.450 & \textbf{0.424} & & LPIPS$\downarrow$ & 0.402 & \textbf{0.382} \\
    \bottomrule[1pt]
    \end{tabular}
}
\label{tab:compregpnerf}

\end{table}

\begin{figure}[t]

\centering
    \includegraphics[width=\linewidth]{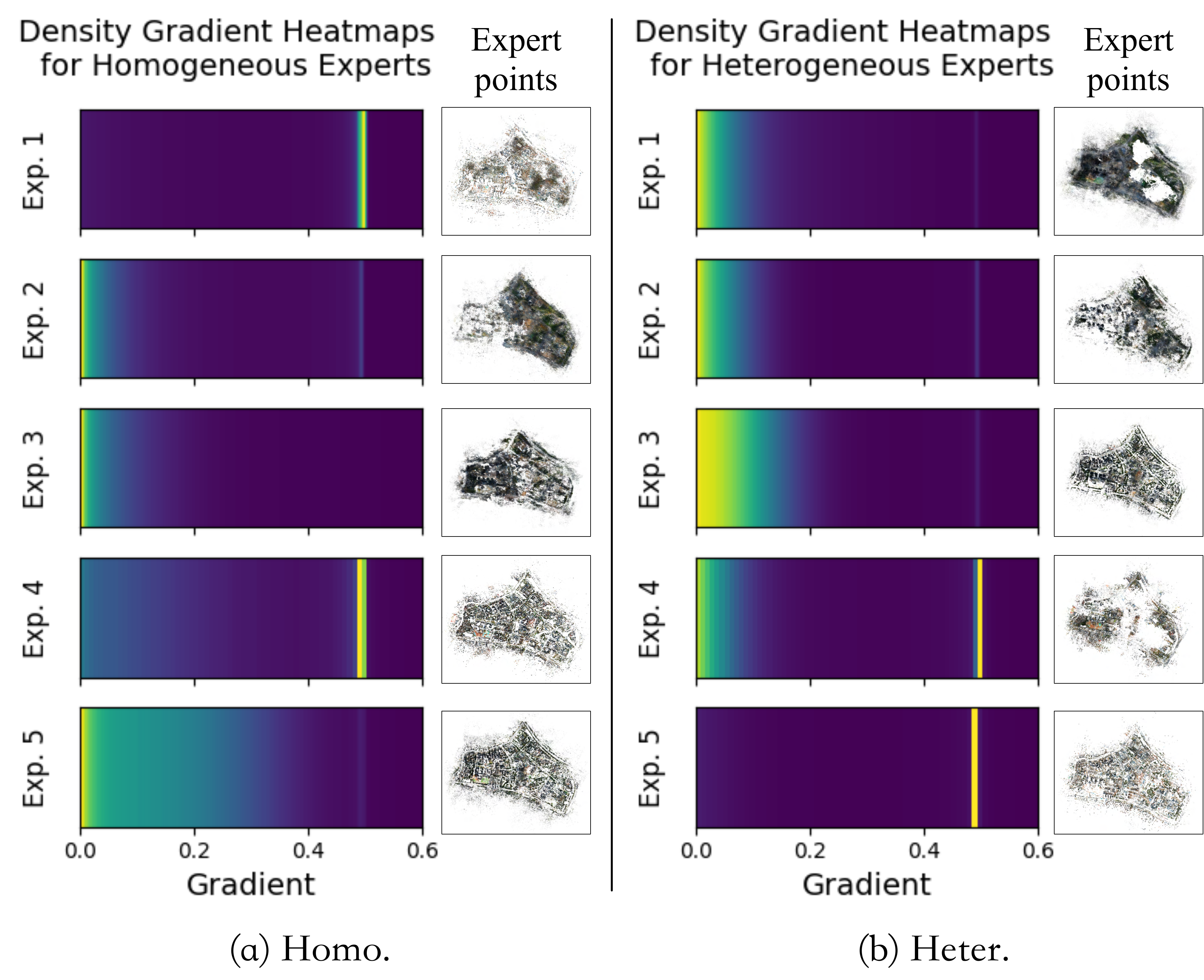}

  \caption{The visualization of the distribution of density gradient and the expert structure for the large UrbanBIS scene. In (a) Homo., the gradient distribution does not show a clear correlation with the experts.  In contrast, in (b) Heter., when the expert resolution increases, the heterogeneous expert handles more points with larger gradients and fewer points with smaller gradients. This gradient analysis also aligns well with the expert point visualization, in which the point clouds in (b) Heter. demonstrate a pattern of increased sharpness along with the increase in expert resolution. This figure illustrates clearly that our heterogeneous structure can learn heterogeneous decomposition and representation.}
  \label{fig:expert_gradient_r1compressed}
\end{figure}

To further analyze the heterogeneous representation learned by heterogeneous experts, we visualize the correlation between the density gradient distribution and the expert structure on the large UrbanBIS scene in Fig.~\ref{fig:expert_gradient_r1compressed}. Our motivation for this analysis is that the density gradient at a 3D point reflects local density variation, indicating the local scene scale. The regions with less density variations can be represented by coarse-scale grids, while the regions with more density variations need fine-grained grids. In Fig.~\ref{fig:expert_gradient_r1compressed}a, the distributions of gradients have no clear correlation with the experts. Different experts tend to cover a wide and homogeneous range of gradients (scales). However, in Fig.~\ref{fig:expert_gradient_r1compressed}b of our heterogeneous model, we can clearly observe that the experts with lower resolutions tend to cover more regions of small gradients, while those with higher resolutions tend to cover more regions of large gradients. This effectively validates that our design can learn heterogeneous decomposition and representation of large-scale scenes.

\begin{table}[htbp]
\centering
\caption{Comparison of the memory and time for testing and training between the efficient hash dispatching in Switch-NeRF++ and MoE dispatching. Our dispatching uses 5 fewer hours and less GPU memory in training, and only half of the time for rendering.
}

\begin{tabular}{c|cc|cc}
  \toprule
        \multirow{2}[0]{*}{Method}  & \multicolumn{2}{c|}{Test} & \multicolumn{2}{c}{Train} \\
          & Mem.$\downarrow$ & Time$\downarrow$ & Mem.$\downarrow$ & Time$\downarrow$      \\
    \midrule
    MoE disp. &  2981M  & 13.5s  &  10386M &  15.2h  \\
    Hash disp. &  \textbf{2857M}  &  \textbf{6.65s} &  \textbf{8707M} &  \textbf{10.7h}   \\
    \bottomrule
    \end{tabular}
\label{table:dispatchingeffiency}
\end{table}

\begin{table}[htbp]
\centering
\caption{Comparison to Instant-NGP~\cite{mueller2022instant} with Mega-NeRF~\cite{Turki_2022_CVPR} (Mega-NGP) and 3D Gaussian spaltting (3D-GS)~\cite{kerbl3Dgaussians} by PSNR. Switch-NeRF++ demonstrates superior accuracy than Mega-NGP and 3D-GS on large-scale scenes.}
\resizebox{\linewidth}{!}{

    \begin{tabular}{c|ccccc}
    \toprule
          & Building & Rubble & Residence & Sci-Art & Campus \\
    \midrule
    3D-GS & 19.04 & 25.06 & 21.37 & 21.65 & 18.74 \\
    Mega-NGP & 20.22 & 24.46 & 21.55 & 24.15 & 22.19 \\
    \midrule
    Switch-NeRF++  & \textbf{21.79} & \textbf{25.55} & \textbf{23.25} & \textbf{26.17} & \textbf{23.94} \\
    \bottomrule
    \end{tabular}%
}
\label{tab:comparegpnerfmegangp}
\end{table}

\noindent \textbf{Gating network design in Switch-NeRF.} We analyze different designs of the MLP-based gating network in Switch-NeRF~\cite{mi2023switchnerf}. Table~\ref{table:decomgatehead-results} compares three configurations. The first is using a Linear gating, which processes the positional encoding $\mbox{PE(\textbf{x})}$ by a trainable linear layer. The second is removing the LayerNorm in the gating network. The third is training the network without the gating loss $L_b$. As shown in the table, Linear gating underperforms due to insufficient network capacity, validating that MLP+Norm design in our gating network can substantially improve reconstruction accuracy. Notably, removing the gating loss $L_b$ prevents the training from converging, highlighting that $L_b$ is necessary for stabilizing the optimization of our sparse MoE system. This finding aligned with prior works ~\cite{DBLP:conf/iclr/ShazeerMMDLHD17, DBLP:conf/iclr/LepikhinLXCFHKS21}.

\noindent \textbf{Full dispatch in Switch-NeRF.} We compare the inference accuracy of using uniform dispatch or using our full dispatch proposed in Sec. \ref{sec:fulldispatch}. The Batch Prioritized Routing~\cite{DBLP:conf/nips/RiquelmePMNJPKH21} and a capacity factor $C_f$ of $2.0$ or $4.0$ are used in the inference of the uniform dispatch. As shown in Table~\ref{table:fulldispatchcompare}, uniform dispatch with $C_f=2.0$ suffers significant quality degradation. While increasing $C_f$ to $4.0$ partially improves the quality, it remains substantially inferior to full dispatch. Meanwhile, a larger $C_f$ introduces extra computational overhead due to zero padding. Visual comparisons in Fig.~\ref{fig:dispatchcompare} further highlight that the uniform dispatch renders images with more artifacts, whereas our full dispatch produces complete and higher-quality outputs. This validates the effectiveness of full dispatch in inference of Switch-NeRF.

\begin{table}[htbp]

  \centering
  \caption{Different gating network designs in Switch-NeRF on Building scene.}

\begin{tabular}{cccc}
  \toprule[1pt]
       & PSNR$\uparrow$ & SSIM$\uparrow$ & LPIPS$\downarrow$ \\
\midrule
 Linear & 20.75 & 0.532 & 0.524 \\
 w/o Norm & 21.39 & 0.577 & \textbf{0.474} \\
 w/o $L_b$ & 15.32 & 0.363 & 0.769 \\
\midrule
 Switch-NeRF & \textbf{21.54} & \textbf{0.579} & \textbf{0.474} \\
\bottomrule[1pt]
\end{tabular}
\label{table:decomgatehead-results}
\end{table}

\begin{table}[htbp]

\centering
\caption{Comparison of Switch-NeRF and Mega-NeRF with different experts on accuracy, test time, test memory, and network parameter numbers. Switch-NeRF demonstrates better scalability than Mega-NeRF, increasing performance without a large increase in test memory and time.
}
\vspace{-8pt}
\resizebox{\linewidth}{!}{\begin{tabular}{ccccccc}
\toprule[1pt]
 Model & PSNR$\uparrow$ & SSIM$\uparrow$ & LPIPS$\downarrow$ & Mem.$\downarrow$& Time$\downarrow$ & Param.$\downarrow$\\
 \midrule
Switch-4 & 21.00 & 0.547 & 0.504 & 5825M & 106s & 2.78M \\
Mega-8 & 20.93 & 0.547 & 0.504 & 6935M & 87.9s & 10.8M \\
Switch-8 & 21.54 & 0.579 & 0.474 & 5847M & 110s & 4.53M \\
Mega-16 & 21.47 & 0.590 & 0.462 & 8042M & 101s & 21.6M \\
Switch-16 & \textbf{22.49} & \textbf{0.625} & \textbf{0.429} & 5876M & 118s & 8.05M \\
\bottomrule[1pt]
\end{tabular}}
\label{table:switchnerfexpertnumber}
\end{table}

\noindent \textbf{Effect of the hash-based gating network.} We replace the hash-based gating network in Switch-NeRF++ with an MLP gating network similar to that in Switch-NeRF~\cite{mi2023switchnerf}, which consists of 4 linear layers with a width of 256. The input of the MLP gating network is the positional encoding~\cite{mildenhall2020nerf}. We train them with 100k steps. As shown in Table~\ref{table:ablationstudy}, the accuracy drops by a large margin with the MLP gating network. This may be because the MLP suffers from slow converging speed, and the representation power of the MLP gating network is inconsistent with the hash experts in Switch-NeRF. This experiment further confirms the importance of our design of the hash-based gating network.

\begin{table}[t]
\centering
\caption{The ablation results of Switch-NeRF using different Top-$1$, Top-$2$ and capacity factors. A larger $C_f$ or employing Top-$2$ operation in the gating network in training can improve the accuracy slightly while consuming more training time and memory.}
\resizebox{\linewidth}{!}{
\begin{tabular}{ccccccc}
\toprule[1pt]
Top-$k$ & Capacity & PSNR$\uparrow$ & SSIM$\uparrow$ & LPIPS$\downarrow$ & Mem.$\downarrow$ & Time$\downarrow$\\
1 & 1.0 & 21.54 & 0.579 & 0.474 & \textbf{10182M} & \textbf{42.5h} \\
1 & 1.5 & 21.70 & \textbf{0.594} & \textbf{0.463} & 12271M & 47.4h \\
2 & 1.0 & \textbf{21.77} & 0.590 & 0.465 & 14548M & 58.7h\\
\bottomrule[1pt]
\end{tabular}}
\label{table:topkfactor}
\end{table}

\begin{table}[t]

\centering
\caption{Ablation results comparing Switch-NeRF++, Switch-NeRF~\cite{mi2023switchnerf}, and Mega-NeRF~\cite{Turki_2022_CVPR} on memory usage and computation time during training and testing. Switch-NeRF++ achieves significantly faster training and testing speeds. Note that Switch-NeRF++ uses a 4x larger batch size than Switch-NeRF and Mega-NeRF. Therefore, it consumes slightly more training memory than Mega-NeRF but still less than Switch-NeRF. In testing, Switch-NeRF++ speed up rendering by \textbf{16x} compared to Switch-NeRF~\cite{mi2023switchnerf} and by \textbf{13x} compared to Mega-NeRF~\cite{Turki_2022_CVPR} with much less memory.}
\resizebox{\linewidth}{!}{
\begin{tabular}{c|cc|cc}
    
  \toprule[1pt]
        \multirow{2}[0]{*}{Method}  & \multicolumn{2}{c|}{Test} & \multicolumn{2}{c}{Train} \\
          & Mem.$\downarrow$ & Time$\downarrow$ & Mem.$\downarrow$ & Time$\downarrow$  \\
    \midrule
    Mega-NeRF & 6935M  & 87.9s & 5124M & 30.7h \\
    Switch-NeRF & 5847M  & 110s & 10182M & 42.5h \\
    Switch-NeRF++ & 2857M & 6.65s & 8707M & 10.7h \\
    \bottomrule[1pt]
    \end{tabular}
    }
\label{table:timeandmemory}
\end{table}

\begin{table}[t]
\centering
\caption{The testing time and accuracy of uniform dispatch with $C_f=2.0$ and $4.0$ and full dispatch. Full dispatch can achieve much better performance with less time usage.}
\begin{tabular}{cccccc}
\toprule[1pt]
Dispatch & PSNR$\uparrow$ & SSIM$\uparrow$ & LPIPS$\downarrow$ & Time (s)$\downarrow$\\
\midrule
Uniform $2.0$ & 17.82 & 0.410 & 0.563 & 131 \\
Uniform $4.0$ & 20.07 & 0.521 & 0.507 & 182 \\
Full & \textbf{21.54} & \textbf{0.579} & \textbf{0.474} & \textbf{110} \\
\bottomrule[1pt]
\end{tabular}
\label{table:fulldispatchcompare}
\end{table}

\begin{table}[htbp]
  \centering
  \caption{Impact of different numbers of gating operations on accuracy and efficiency in Switch-NeRF. Despite maintaining the same number of expert layers, Switch-NeRF with two gating operations fails to enhance rendering accuracy compared to a single gating operation, while requiring more memory usage and longer training and inference times.}
\resizebox{\linewidth}{!}{
  
    \begin{tabular}{c|ccc|cc|cc}
    \toprule
    \multicolumn{1}{c|}{\multirow{2}[2]{*}{Gating number}} & \multicolumn{1}{c}{\multirow{2}[2]{*}{PSNR$\uparrow$}} & \multicolumn{1}{c}{\multirow{2}[2]{*}{SSIM$\uparrow$}} & \multicolumn{1}{c|}{\multirow{2}[2]{*}{LPIPS$\downarrow$}} & \multicolumn{2}{c|}{Train} & \multicolumn{2}{c}{Test} \\
          &       &       &       & Mem.   & Time  & Mem.   & Time \\
    \midrule
    1     & 21.54 & 0.579 & 0.474 & 10182M & 42.5h & 5847M & 110s \\
    2     & 21.55 & 0.574 & 0.477 & 15315M & 56.2h & 5838M & 152s \\
    \bottomrule
    \end{tabular}
    }
  \label{tab:2moelayers}%
\end{table}%

\begin{figure}
\centering
\includegraphics[width=\linewidth]{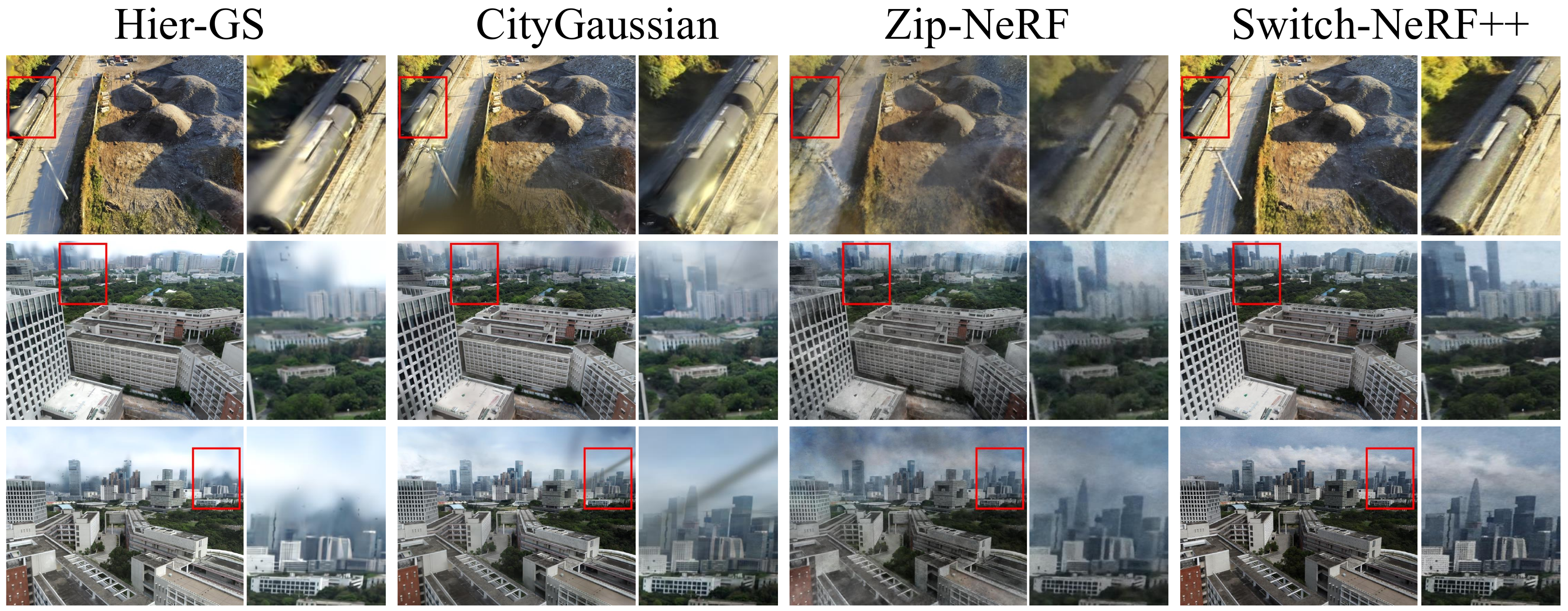}
  \caption{Comparison with large-scale Gaussian methods and Zip-NeRF. The zoomed-in image patches show that GS-based methods such as CityGaussian and Hier-GS tend to produce floating artifacts and blurred structures in challenging regions, while Switch-NeRF++ can produce high-fidelity and stable reconstructions.}
  \label{fig:gs_visualization}
\end{figure}

\noindent \textbf{Effect of the number of experts.} 
We analyze the effect of the number of NeRF experts in both Switch-NeRF~\cite{mi2023switchnerf}.
Table~\ref{table:switchnerfexpertnumber} reveals that even using 4 experts, Switch-NeRF already matches the accuracy of Mega-NeRF with 8 experts. Scaling Switch-NeRF to 16 experts yields significant performance gains compared to 8 experts, while incurring minimal additional computational and memory overhead during inference. This contrasts with the Mega-NeRF with 16 experts, where Switch-NeRF not only scales much better and achieves a higher accuracy, but also reduces parameters and memory usage. The results demonstrate the superior scalability of Switch-NeRF in balancing model capacity and resource demands.

\noindent \textbf{Effect of Top-1 and Top-2 gating.} In Switch-NeRF~\cite{mi2023switchnerf} and Switch-NeRF++, we only choose one expert for a 3D scene point to perform a scene decomposition. However, similar to MoE methods~\cite{DBLP:conf/iclr/ShazeerMMDLHD17, mi2023switchnerf}, we can also use a Top-$2$ operation to select 2 experts for each sample point. The final expert output is the weighted sum of the two expert outputs by the predicted gating values. As shown in Table~\ref{table:ablationstudy} and Table~\ref{table:topkfactor}, using 2 experts can achieve comparable results to the Top-$1$ selection. This shows that using only one expert for each sample is already sufficient for good results. Using more experts for a sample without increasing the network capacity does not necessarily produce better results.

\begin{table*}[htbp]
  \centering
  \caption{Comparison with large-scale GS methods and Zip-NeRF. Experiment settings are aligned with VastGaussian.}
  \resizebox{\linewidth}{!}{
    \begin{tabular}{c|ccc|ccc|ccc|ccc|ccc}
    \toprule
    Dataset & \multicolumn{3}{c|}{Building} & \multicolumn{3}{c|}{Rubble} & \multicolumn{3}{c|}{Residence} & \multicolumn{3}{c|}{Sci-Art} & \multicolumn{3}{c}{Campus} \\
    \midrule
    Metrics & PSNR  & SSIM  & LPIPS & PSNR  & SSIM  & LPIPS & PSNR  & SSIM  & LPIPS & PSNR  & SSIM  & LPIPS & PSNR  & SSIM  & LPIPS \\
    \midrule
    Hier-GS & 22.24 & 0.711 & 0.217 & 24.97 & 0.715 & 0.226    & 23.27 & 0.775 & 0.175 & 26.60 & 0.857 & 0.155 & -- & -- & -- \\

    Zip-NeRF & 16.22 & 0.418 & 0.548 & 21.37 & 0.526 & 0.474 & 19.75 & 0.598 & 0.383 & 19.41 & 0.635 & 0.399 & 16.60 & 0.406 & 0.698 \\
    VastGaussian & 23.50 & 0.804 & 0.130 & 26.92 & 0.823 & 0.132 & 24.25 & 0.852 & 0.124 & 26.81 & 0.885 & 0.121 & 26.00 & 0.816 & 0.151 \\
    CityGaussian & 22.43 & 0.761 & 0.179 & 26.08 & 0.793 & 0.164 & 23.12 & 0.809 & 0.161 & 24.42 & 0.837 & 0.171 & -- & -- & -- \\
    \midrule
    Switch-NeRF++ & 22.41 & 0.633 & 0.290 & 26.02 & 0.680 & 0.281 & 24.00 & 0.730 & 0.224 & 27.40 & 0.823 & 0.190 & 24.37 & 0.594 & 0.429 \\

    \bottomrule
    \end{tabular}%
    }
  \label{tab:LargescaleGaussiansplatting}%
\end{table*}%

\begin{figure}[htbp]
    \centering

    \centering
    \includegraphics[width=\linewidth]{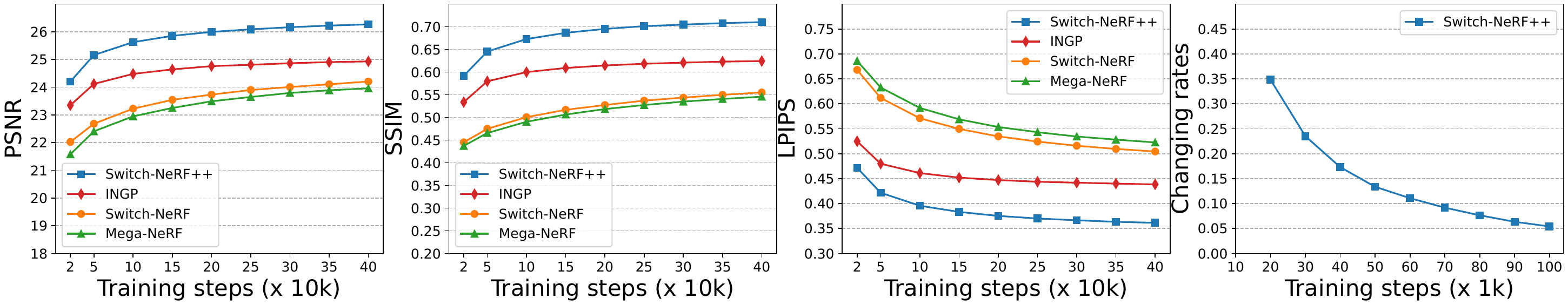}
    \caption{The accuracy of our Switch-NeRF++ and other methods with respect to training steps. Our method continues to improve with more training steps. We also calculate the changing rates of expert assignment with respect to the training steps. In the early training, the changing rate is at a reasonably high value to enable the network to optimize the gate assignment. The changing rate decreases to about 0.05, showing that the gating network has converged. Note that the steps are x1k in the plot.}
    \label{fig:mohe_switch_mega_ingp_all}
\end{figure}

\noindent \textbf{Effect of training steps.} Fig.~\ref{fig:mohe_switch_mega_ingp_all} shows the PSNR, SSIM, and LPIPS of Switch-NeRF and Switch-NeRF++ compared to other methods on the Rubble scene in different training steps. Our Switch-NeRF++ can already obtain state-of-the-art results with 50k-100k steps and clearly outperform Switch-NeRF. The accuracy can continue to improve a lot with more training steps. This means that given more computation, Switch-NeRF++ can produce better results, which enables our network to fit flexibly into different application scenarios, and we can flexibly make a good trade-off between the accuracy and the computation overhead.

\begin{table}[t]
  \centering
  \caption{Additional ablation studies on the detailed configurations of Switch-NeRF++, evaluated using the PSNR metric (higher is better). \textbf{Gumbel} uses Gumbel-Softmax for expert selection. \textbf{E16} denotes 16 experts; \textbf{HS2} uses 8 experts with a doubled hash table size. \textbf{E8,HS1/8} refers to 8 experts with 1/8 of the original hash table size, while \textbf{INGP} is an Instant-NGP baseline using a single hash grid with the same number of parameters as E8,HS1/8. \textbf{HS1/2,F2} indicates half the hash table size with doubled feature dimensions. \textbf{Head} uses the same prediction head as Switch-NeRF. \textbf{Ours} shows the performance of our standard Switch-NeRF++ settings.}
  \resizebox{\linewidth}{!}{
    \begin{tabular}{ccccccc|c}
    \toprule
    Gumbel & E16 & HS2 & E8,HS1/8 & INGP & HS1/2,F2 & Head  & Ours \\
    \midrule
    18.22 & 23.55 & 23.35 & 22.32 & 21.41 & 23.22 &   23.31    & 23.25 \\
    \toprule
    \end{tabular}%
    }
  \label{tab:Moreablationstudies}%
\end{table}%

\noindent \textbf{Plane features without decomposition.} The GP-NeRF~\cite{zhang2023efficient} augments hash encodings with 2D ground plane features. It does not perform scene decomposition. The comparison results are shown in Table~\ref{tab:compregpnerf}. Switch-NeRF++ consistently outperforms GP-NeRF, especially on the very large UrbanBIS~\cite{UrbanBISbib} dataset. These experiments validate the importance of our learnable heterogeneous decomposition on large-scale scenes to better exploit network parameters.

\noindent \textbf{Manual decomposition.}
In Table \ref{tab:comparegpnerfmegangp}, we integrate INGP into the manual decomposition method Mega-NeRF~\cite{Turki_2022_CVPR} as Mega-NGP. It manually divides the scene into multiple parts according to 3D distances. Each part is handled by a typical INGP network. Switch-NeRF++ significantly outperforms Mega-NGP, indicating that the spatial decomposition in Mega-NeRF is not well-suited for hash grids. In contrast, our method is designed to learn the heterogeneous decomposition and representation of large-scale scenes, which has been proven more generally applicable and accurate.

\noindent \textbf{Vanilla Gaussian splatting.} We compare with the vanilla Gaussian splatting method (3D-GS)~\cite{kerbl3Dgaussians} in Table~\ref{tab:comparegpnerfmegangp} on large-scale datasets. The results demonstrate that the vanilla 3D-GS struggles on large-scale scenes. It relies on a point cloud for initialization, while point clouds in large-scale scenes are typically noisy, contain outliers, and have significantly varying densities. The original 3D-GS has no special designs for these problems, resulting in poor initialization and training. Our method incorporates the learnable heterogeneous decomposition specially designed for large-scale scenes and can get much better results. Moreover, our method does not need any point clouds for initialization, making it more suitable for various large-scale reconstruction tasks.

\begin{table}[htbp]
  \centering
  \caption{A detailed analysis of changes in memory usage and computational complexity with network designs in Switch-NeRF++. The training batch size and steps are aligned with the Switch-NeRF++ in our main results.}
  \resizebox{\linewidth}{!}{
    \begin{tabular}{ccc|cccc}
    \toprule

    \multirow{2}[0]{*}{Hash Gate} & \multirow{2}[0]{*}{Effic. Disp.} & \multirow{2}[0]{*}{Hash Exp. \& Head} & \multicolumn{2}{c}{Train} & \multicolumn{2}{c}{Test} \\
          &       &       & Mem.  & Time  & Mem.  & Time \\
    \midrule
      $\checkmark$ &       &       & 18503M & 27.4h & 5461M & 104.2s \\
     $\checkmark$  &  $\checkmark$  &       & 15859M & 22.0h & 4430M & 71.4s \\
      $\checkmark$  &  $\checkmark$   &  $\checkmark$  & 8707M & 10.7h & 2857M & 6.65s \\
    \bottomrule
    \end{tabular}%
    }
  \label{tab:analysisofchanges}%
\end{table}%

\noindent \textbf{Large-scale Gaussian splatting and Zip-NeRF.} We compare with Zip-NeRF~\cite{barron2023zip} and Large-scale Gaussian splatting methods such as Hierarchical-3DGaussians~\cite{hierarchicalgaussians24}, CityGaussian~\cite{liu2024citygaussian}, and Vast-Gaussian~\cite{lin2024vastgaussian} in Table~\ref{tab:LargescaleGaussiansplatting}. In these experiments, we align the training and evaluation settings of Switch-NeRF++ with Vast-Gaussian, such as the training and evaluation resolutions, the color correction in evaluation and the AlexNet version of LPIPS. We include results for Hierarchical-3DGaussians and Zip-NeRF using their open-source training and evaluation codes. We obtain the results of City-Gaussian using its released checkpoints. All evaluation metrics of these methods are also aligned with the Vast-Gaussian setting.

As shown in Table~\ref{tab:LargescaleGaussiansplatting}, Switch-NeRF++ achieves significantly higher PSNR than Hierarchical-3DGaussians. Hierarchical-3DGaussians, which decomposes the scene via a tree structure and merges chunks by post-training, struggles to generalize across different scene types due to its rule-based design. Compared with CityGaussian, Switch-NeRF++ achieves comparable PSNR on Building and Rubble and performs notably better on Residence and Sci-Art (27.40 vs. 24.42). This suggests that rule-based large-scale Gaussian methods still face challenges in handling sparse supervision, noisy point clouds, and diverse scene types, which are common issues in large-scale settings. While Switch-NeRF++ underperforms Vast-Gaussian on some scenes (e.g., Building, Rubble, and Residence), it performs competitively and surpasses it on Sci-Art, which contains challenging, distant regions with sparse point clouds.

Although GS methods generally report better SSIM and LPIPS, this is partly because of their use of SSIM loss. This may also stem from the fundamental differences between GS-based rasterization and our NeRF-style ray-based rendering. GS-based methods are fundamentally point-based rendering based on sparse point clouds, which may focus more on sharp structures and may get higher SSIM scores. Our NeRF-style ray-based rendering may put more emphasis on global consistency and smoothness among 3D points. However, Fig.~\ref{fig:gs_visualization} shows that GS-based methods tend to produce visible artifacts in air or distant regions, which may be overlooked by SSIM. Our method can produce much cleaner and more globally consistent images.

While Zip-NeRF~\cite{barron2023zip} also extends Instant-NGP for multi-scale rendering, it is not specifically designed for large-scale scenes. As shown in Table~\ref{tab:LargescaleGaussiansplatting} and Fig.~\ref{fig:gs_visualization} (Paper Fig. 12), Zip-NeRF struggles in large-scale settings, whereas Switch-NeRF++ achieves superior results across various scenes.

Fig.~\ref{fig:gs_visualization} presents a qualitative comparison between Switch-NeRF++ and large-scale Gaussian methods. GS-based methods often suffer from artifacts in distant or low-density regions due to their reliance on high-quality input point clouds. In contrast, our method maintains robustness across these difficult areas.

These limitations of GS-based methods arise from their reliance on high-quality point clouds and dense image coverage. However, these conditions are rarely met in large-scale scene datasets, which typically contain sparse images considering the large areas they cover, and noisy and sparse point clouds. Additionally, the rule-based decomposition used in GS-based methods struggles to generalize across different scene types. As a result, large-scale GS methods still face challenges in these settings, whereas Switch-NeRF++ remains robust and competitive.

\noindent \textbf{More ablation studies.} Table~\ref{tab:Moreablationstudies} presents additional ablations on the Residence scene. Using Gumbel-Softmax for expert selection leads to unstable training and poor results, confirming the effectiveness of the MoE structure. Increasing the number of experts (E16) yields greater accuracy gains than enlarging the hash map (HS2), likely due to improved specialization and better hash table utilization. In contrast, larger hash maps may result in under-utilized entries. Reducing the hash map size while doubling the feature dimension (``HS1/2,F2") to maintain parameter count shows minimal accuracy loss, indicating robustness to higher collision rates. Replacing the prediction head in Switch-NeRF++ with that of Switch-NeRF has little impact, suggesting that gains stem mainly from the heterogeneous hash expert design.

\subsection{Efficiency Analysis}

\textbf{Efficiency of Switch-NeRF.} Table~\ref{table:timeandmemory} compares the efficiency of Switch-NeRF~\cite{mi2023switchnerf} and Mega-NeRF. With only \emph{half} network parameters, Switch-NeRF attains superior accuracy than Mega-NeRF. Each expert in Mega-NeRF employs a foreground and a background NeRF and a set of appearance embeddings (AE), which results in more network parameters. In contrast, Switch-NeRF designs an end-to-end framework that needs only one background NeRF and one set of AE for the whole scene, largely decreasing the parameters while getting better accuracy. Since Switch-NeRF has an extra gating network compared to Mega-NeRF, it uses slightly more floating point operations (FLOPs) for each point, resulting in slightly more memory and time usage for training. However, Switch-NeRF consumes around 20\% less testing memory than Mega-NeRF while only increasing minor testing time. These results demonstrate that Switch-NeRF can achieve a much better accuracy with a large influence on the efficiency, validating the efficiency and wide applicability of our sparse MoE structure on large-scale scene modeling.

\noindent \textbf{Efficient hash dispatching.} We analyze the time and memory efficiency of the hash dispatching of Switch-NeRF++ compared to the optimized dispatching in the MoE method~\cite{hwang2022tutel, mi2023switchnerf} in Switch-NeRF. As shown in Table~\ref{table:dispatchingeffiency}, when trained with 100k steps, the MoE dispatching uses 5 more hours and more GPU memory. When rendering images, our method uses only 6.65s to render a $1152\times 864$ image, while the MoE dispatching costs 13.5s. These results show that our efficient dispatching can save time and memory for both training and testing by a large margin compared to the MoE dispatching. The memory is calculated by the Pytorch APIs.

\begin{table}[htbp]
  \centering
  \caption{Comparison of efficiency with GS-based methods on Building scene. Our method remains competitive in training efficiency on large-scale scenes. While GS-based methods can render efficiently, our model is significantly more compact and parameter-efficient, due to compact hash experts and the sparse MoE network architecture.}
    \resizebox{\linewidth}{!}{
    \begin{tabular}{c|cccc}
    \toprule
          & Recon. time & Rend. time & Rend. mem. & Model size \\
    \midrule
    Hier-GS &   3.20h    & 0.223s & 18147M & 855M \\
    CityGaussian &   7.25h    & 0.031s & 8985M & 787M \\
    \midrule
    Switch-NeRF++  & 10.7h & 6.650s & 2857M & 119M \\
    \bottomrule
    \end{tabular}%
    }
  \label{tab:Efficiencygs}%
\end{table}%

\noindent \textbf{Time and memory overhead for rendering of Switch-NeRF++.} In this experiment, we evaluate the time and memory overhead of Switch-NeRF++ for rendering images. As highlighted in Table~\ref{table:timeandmemory}, our Switch-NeRF++ can render a $1152\times 864$ image in only 6.65s, which is a 16x acceleration compared to Switch-NeRF and a 13x acceleration compared to Mega-NeRF. The memory used by our method is even less than half of these two methods. This experiment clearly verifies the superior efficiency of our method. The efficiency enables our method to be applied to real-world, challenging, and large-scale scenes.

\noindent \textbf{Breakdown analysis of efficiency.}
To better substantiate the efficiency of Switch-NeRF++, we present a detailed breakdown of memory and time usage during training and testing by adding our components to the original Switch-NeRF. The training batch size and steps are aligned with Switch-NeRF++ in our main results in Table~\ref{table:magadataresults}. As shown in Table~\ref{tab:analysisofchanges}, using efficient hash gating, dispatching, and hash-based experts leads to substantial improvements. Specifically, replacing large MLP experts with efficient hash experts dramatically reduces memory usage and accelerates both training and inference. This is because most of the computation overhead of Switch-NeRF is in its MLP experts.

\noindent \textbf{Efficiency comparison with GS-based methods.} Table~\ref{tab:Efficiencygs} reports the comparison of reconstruction time, rendering time, memory usage, and model size on the Building scene with GS-based methods. Our model achieves significantly smaller memory cost and model size, due to its compact hash experts and sparse MoE structure. CityGaussian takes 7.25 hours on 4 GPUs (a coarse stage plus 20 chunks) for training, indicating that Switch-NeRF++ remains competitively efficient in training. While GS-based methods excel in rendering speed, they face scalability and robustness challenges on large-scale scenes with sparse images and noisy point clouds. Switch-NeRF++ provides a more compact, robust, and general solution for large-scale scene modeling.

\section{Conclusion}
We propose Switch-NeRF++, a novel framework for highly scalable Neural Radiance Field (NeRF) modeling. It simultaneously addresses the challenges of learnable decomposition, heterogeneity, and efficiency in large-scale NeRF by incorporating the Heterogeneous Mixture of Hash Experts (HMoHE) architecture. Extensive experiments show the superior accuracy and scalability of our method compared to the best-performing methods. Our framework emerges as a practical solution for applying NeRF in real-world large-scale scenes with flexible scalability and high efficiency.

\ifCLASSOPTIONcompsoc

\ifCLASSOPTIONcaptionsoff
  \newpage
\fi

\bibliographystyle{ieeetr}
\bibliography{tpami}

\begin{thebibliography}{10}

\bibitem{mildenhall2020nerf}
B.~Mildenhall, P.~P. Srinivasan, M.~Tancik, J.~T. Barron, R.~Ramamoorthi, and R.~Ng, ``Nerf: Representing scenes as neural radiance fields for view synthesis,'' in {\em ECCV}, 2020.

\bibitem{Turki_2022_CVPR}
H.~Turki, D.~Ramanan, and M.~Satyanarayanan, ``Mega-nerf: Scalable construction of large-scale nerfs for virtual fly-throughs,'' in {\em CVPR}, 2022.

\bibitem{Tancik_2022_CVPR}
M.~Tancik, V.~Casser, X.~Yan, S.~Pradhan, B.~Mildenhall, P.~P. Srinivasan, J.~T. Barron, and H.~Kretzschmar, ``Block-nerf: Scalable large scene neural view synthesis,'' in {\em CVPR}, 2022.

\bibitem{mi2023switchnerf}
Z.~MI and D.~Xu, ``Switch-ne{RF}: Learning scene decomposition with mixture of experts for large-scale neural radiance fields,'' in {\em ICLR}, 2023.

\bibitem{mueller2022instant}
T.~M\"uller, A.~Evans, C.~Schied, and A.~Keller, ``Instant neural graphics primitives with a multiresolution hash encoding,'' {\em ACM Trans. Graph.}, vol.~41, no.~4, pp.~102:1--102:15, 2022.

\bibitem{SunSC22}
C.~Sun, M.~Sun, and H.~Chen, ``Direct voxel grid optimization: Super-fast convergence for radiance fields reconstruction,'' in {\em CVPR}, 2022.

\bibitem{yu_and_fridovichkeil2021plenoxels}
S.~Fridovich-Keil, A.~Yu, M.~Tancik, Q.~Chen, B.~Recht, and A.~Kanazawa, ``Plenoxels: Radiance fields without neural networks,'' in {\em CVPR}, 2022.

\bibitem{Chen2022ECCV}
A.~Chen, Z.~Xu, A.~Geiger, J.~Yu, and H.~Su, ``Tensorf: Tensorial radiance fields,'' in {\em ECCV}, 2022.

\bibitem{xu2023gridguided}
L.~Xu, Y.~Xiangli, S.~Peng, X.~Pan, N.~Zhao, C.~Theobalt, B.~Dai, and D.~Lin, ``Grid-guided neural radiance fields for large urban scenes,'' in {\em CVPR}, 2023.

\bibitem{UrbanBISbib}
G.~Yang, F.~Xue, Q.~Zhang, K.~Xie, C.-W. Fu, and H.~Huang, ``Urbanbis: a large-scale benchmark for fine-grained urban building instance segmentation,'' in {\em SIGGRAPH}, 2023.

\bibitem{DBLP:conf/iclr/ShazeerMMDLHD17}
N.~Shazeer, A.~Mirhoseini, K.~Maziarz, A.~Davis, Q.~V. Le, G.~E. Hinton, and J.~Dean, ``Outrageously large neural networks: The sparsely-gated mixture-of-experts layer,'' in {\em ICLR}, 2017.

\bibitem{fedus2022switch}
W.~Fedus, B.~Zoph, and N.~Shazeer, ``Switch transformers: Scaling to trillion parameter models with simple and efficient sparsity,'' {\em JMLR}, 2022.

\bibitem{lowe1999object}
D.~G. Lowe, ``Object recognition from local scale-invariant features,'' in {\em ICCV}, vol.~2, pp.~1150--1157, IEEE, 1999.

\bibitem{cech2010efficient}
J.~Cech, J.~Matas, and M.~Perdoch, ``Efficient sequential correspondence selection by cosegmentation,'' {\em TPAMI}, vol.~32, no.~9, pp.~1568--1581, 2010.

\bibitem{schonberger2017vote}
J.~L. Sch{\"o}nberger, T.~Price, T.~Sattler, J.-M. Frahm, and M.~Pollefeys, ``A vote-and-verify strategy for fast spatial verification in image retrieval,'' in {\em ACCV}, pp.~321--337, Springer, 2017.

\bibitem{sarlin2020superglue}
P.-E. Sarlin, D.~DeTone, T.~Malisiewicz, and A.~Rabinovich, ``Superglue: Learning feature matching with graph neural networks,'' in {\em CVPR}, pp.~4938--4947, 2020.

\bibitem{snavely2006photo}
N.~Snavely, S.~M. Seitz, and R.~Szeliski, ``Photo tourism: exploring photo collections in 3d,'' in {\em SIGGRAPH}, pp.~835--846, 2006.

\bibitem{crandall2011discrete}
D.~Crandall, A.~Owens, N.~Snavely, and D.~Huttenlocher, ``Discrete-continuous optimization for large-scale structure from motion,'' in {\em CVPR}, pp.~3001--3008, IEEE, 2011.

\bibitem{schonberger2016structure}
J.~L. Schonberger and J.-M. Frahm, ``Structure-from-motion revisited,'' in {\em CVPR}, pp.~4104--4113, 2016.

\bibitem{furukawa2009accurate}
Y.~Furukawa and J.~Ponce, ``Accurate, dense, and robust multiview stereopsis,'' {\em IEEE TPAMI}, vol.~32, no.~8, pp.~1362--1376, 2009.

\bibitem{schonberger2016pixelwise}
J.~L. Sch{\"o}nberger, E.~Zheng, J.-M. Frahm, and M.~Pollefeys, ``Pixelwise view selection for unstructured multi-view stereo,'' in {\em ECCV}, pp.~501--518, Springer, 2016.

\bibitem{xu2019multi}
Q.~Xu and W.~Tao, ``Multi-scale geometric consistency guided multi-view stereo,'' in {\em CVPR}, pp.~5483--5492, 2019.

\bibitem{mi2022generalized}
Z.~Mi, C.~Di, and D.~Xu, ``Generalized binary search network for highly-efficient multi-view stereo,'' in {\em CVPR}, pp.~12991--13000, 2022.

\bibitem{yao2018mvsnet}
Y.~Yao, Z.~Luo, S.~Li, T.~Fang, and L.~Quan, ``Mvsnet: Depth inference for unstructured multi-view stereo,'' in {\em ECCV}, pp.~767--783, 2018.

\bibitem{chen2019point}
R.~Chen, S.~Han, J.~Xu, and H.~Su, ``Point-based multi-view stereo network,'' in {\em ICCV}, pp.~1538--1547, 2019.

\bibitem{gu2020cascade}
X.~Gu, Z.~Fan, S.~Zhu, Z.~Dai, F.~Tan, and P.~Tan, ``Cascade cost volume for high-resolution multi-view stereo and stereo matching,'' in {\em CVPR}, pp.~2495--2504, 2020.

\bibitem{xu2020learning}
Q.~Xu and W.~Tao, ``Learning inverse depth regression for multi-view stereo with correlation cost volume,'' in {\em AAAI}, vol.~34, pp.~12508--12515, 2020.

\bibitem{zhang2020nerf++}
K.~Zhang, G.~Riegler, N.~Snavely, and V.~Koltun, ``Nerf++: Analyzing and improving neural radiance fields,'' {\em arXiv preprint arXiv:2010.07492}, 2020.

\bibitem{martin2021nerf}
R.~Martin-Brualla, N.~Radwan, M.~S. Sajjadi, J.~T. Barron, A.~Dosovitskiy, and D.~Duckworth, ``Nerf in the wild: Neural radiance fields for unconstrained photo collections,'' in {\em CVPR}, 2021.

\bibitem{barron2021mip}
J.~T. Barron, B.~Mildenhall, M.~Tancik, P.~Hedman, R.~Martin-Brualla, and P.~P. Srinivasan, ``Mip-nerf: A multiscale representation for anti-aliasing neural radiance fields,'' in {\em ICCV}, 2021.

\bibitem{barron2022mip}
J.~T. Barron, B.~Mildenhall, D.~Verbin, P.~P. Srinivasan, and P.~Hedman, ``Mip-nerf 360: Unbounded anti-aliased neural radiance fields,'' in {\em CVPR}, pp.~5470--5479, 2022.

\bibitem{kangle2021dsnerf}
K.~Deng, A.~Liu, J.-Y. Zhu, and D.~Ramanan, ``Depth-supervised {NeRF}: Fewer views and faster training for free,'' in {\em CVPR}, June 2022.

\bibitem{sparf2023}
P.~Truong, M.-J. Rakotosaona, F.~Manhardt, and F.~Tombari, ``Sparf: Neural radiance fields from sparse and noisy poses,'' CVPR, 2023.

\bibitem{zhong2024cvt}
Y.~Zhong, L.~Hong, Z.~Li, and D.~Xu, ``Cvt-xrf: Contrastive in-voxel transformer for 3d consistent radiance fields from sparse inputs,'' in {\em CVPR}, 2024.

\bibitem{wang2021neus}
P.~Wang, L.~Liu, Y.~Liu, C.~Theobalt, T.~Komura, and W.~Wang, ``Neus: Learning neural implicit surfaces by volume rendering for multi-view reconstruction,'' {\em NeurIPS}, 2021.

\bibitem{li2023neuralangelo}
Z.~Li, T.~M\"uller, A.~Evans, R.~H. Taylor, M.~Unberath, M.-Y. Liu, and C.-H. Lin, ``Neuralangelo: High-fidelity neural surface reconstruction,'' in {\em CVPR}, 2023.

\bibitem{schoenberger2016sfm}
J.~L. Sch\"{o}nberger and J.-M. Frahm, ``Structure-from-motion revisited,'' in {\em CVPR}, 2016.

\bibitem{wang2021nerf}
Z.~Wang, S.~Wu, W.~Xie, M.~Chen, and V.~A. Prisacariu, ``Nerf--: Neural radiance fields without known camera parameters,'' {\em arXiv preprint arXiv:2102.07064}, 2021.

\bibitem{lin2021barf}
C.-H. Lin, W.-C. Ma, A.~Torralba, and S.~Lucey, ``Barf: Bundle-adjusting neural radiance fields,'' in {\em IEEE International Conference on Computer Vision ({ICCV})}, 2021.

\bibitem{yu2021plenoctrees}
A.~Yu, R.~Li, M.~Tancik, H.~Li, R.~Ng, and A.~Kanazawa, ``{PlenOctrees} for real-time rendering of neural radiance fields,'' in {\em ICCV}, 2021.

\bibitem{garbin2021fastnerf}
S.~J. Garbin, M.~Kowalski, M.~Johnson, J.~Shotton, and J.~Valentin, ``Fastnerf: High-fidelity neural rendering at 200fps,'' in {\em ICCV}, 2021.

\bibitem{Reiser_2021_ICCV}
C.~Reiser, S.~Peng, Y.~Liao, and A.~Geiger, ``Kilonerf: Speeding up neural radiance fields with thousands of tiny mlps,'' in {\em ICCV}, 2021.

\bibitem{Fridovich_Keil_2022_CVPR}
S.~Fridovich-Keil, A.~Yu, M.~Tancik, Q.~Chen, B.~Recht, and A.~Kanazawa, ``Plenoxels: Radiance fields without neural networks,'' in {\em CVPR}, pp.~5501--5510, June 2022.

\bibitem{kerbl3Dgaussians}
B.~Kerbl, G.~Kopanas, T.~Leimk{\"u}hler, and G.~Drettakis, ``3d gaussian splatting for real-time radiance field rendering,'' {\em ACM Transactions on Graphics}, 2023.

\bibitem{yu2021pixelnerf}
A.~Yu, V.~Ye, M.~Tancik, and A.~Kanazawa, ``pixelnerf: Neural radiance fields from one or few images,'' in {\em CVPR}, pp.~4578--4587, 2021.

\bibitem{chen2021mvsnerf}
A.~Chen, Z.~Xu, F.~Zhao, X.~Zhang, F.~Xiang, J.~Yu, and H.~Su, ``Mvsnerf: Fast generalizable radiance field reconstruction from multi-view stereo,'' in {\em Proceedings of the IEEE/CVF international conference on computer vision}, pp.~14124--14133, 2021.

\bibitem{wei2021nerfingmvs}
Y.~Wei, S.~Liu, Y.~Rao, W.~Zhao, J.~Lu, and J.~Zhou, ``Nerfingmvs: Guided optimization of neural radiance fields for indoor multi-view stereo,'' in {\em ICCV}, 2021.

\bibitem{wang2021ibrnet}
Q.~Wang, Z.~Wang, K.~Genova, P.~P. Srinivasan, H.~Zhou, J.~T. Barron, R.~Martin-Brualla, N.~Snavely, and T.~Funkhouser, ``Ibrnet: Learning multi-view image-based rendering,'' in {\em CVPR}, 2021.

\bibitem{chan2022efficient}
E.~R. Chan, C.~Z. Lin, M.~A. Chan, K.~Nagano, B.~Pan, S.~De~Mello, O.~Gallo, L.~J. Guibas, J.~Tremblay, S.~Khamis, {\em et~al.}, ``Efficient geometry-aware 3d generative adversarial networks,'' in {\em CVPR}, pp.~16123--16133, 2022.

\bibitem{honglrm}
Y.~Hong, K.~Zhang, J.~Gu, S.~Bi, Y.~Zhou, D.~Liu, F.~Liu, K.~Sunkavalli, T.~Bui, and H.~Tan, ``Lrm: Large reconstruction model for single image to 3d,'' in {\em ICLR}, 2024.

\bibitem{dong2024interactive3d}
S.~Dong, L.~Ding, Z.~Huang, Z.~Wang, T.~Xue, and D.~Xu, ``Interactive3d: Create what you want by interactive 3d generation,'' in {\em Proceedings of the IEEE/CVF Conference on Computer Vision and Pattern Recognition}, pp.~4999--5008, 2024.

\bibitem{spinnerf}
A.~Mirzaei, T.~Aumentado-Armstrong, K.~G. Derpanis, J.~Kelly, M.~A. Brubaker, I.~Gilitschenski, and A.~Levinshtein, ``{SPIn-NeRF}: Multiview segmentation and perceptual inpainting with neural radiance fields,'' in {\em CVPR}, 2023.

\bibitem{wang2024gscream}
Y.~Wang, Q.~Wu, G.~Zhang, and D.~Xu, ``Gscream: Learning 3d geometry and feature consistent gaussian splatting for object removal,'' in {\em ECCV}, 2024.

\bibitem{zhang2023efficient}
Y.~Zhang, G.~Chen, and S.~Cui, ``Efficient large-scale scene representation with a hybrid of high-resolution grid and plane features,'' {\em arXiv preprint arXiv:2303.03003}, 2023.

\bibitem{kerbl20233d}
B.~Kerbl, G.~Kopanas, T.~Leimk{\"u}hler, and G.~Drettakis, ``3d gaussian splatting for real-time radiance field rendering,'' {\em ToG}, pp.~1--14, 2023.

\bibitem{hierarchicalgaussians24}
B.~Kerbl, A.~Meuleman, G.~Kopanas, M.~Wimmer, A.~Lanvin, and G.~Drettakis, ``A hierarchical 3d gaussian representation for real-time rendering of very large datasets,'' {\em ACM ToG}, vol.~43, no.~4, 2024.

\bibitem{liu2024citygaussian}
Y.~Liu, H.~Guan, C.~Luo, L.~Fan, N.~Wang, J.~Peng, and Z.~Zhang, ``Citygaussian: Real-time high-quality large-scale scene rendering with gaussians,'' {\em arXiv preprint arXiv:2404.01133}, 2024.

\bibitem{chen2024dogaussian}
Y.~Chen and G.~H. Lee, ``Dogaussian: Distributed-oriented gaussian splatting for large-scale 3d reconstruction via gaussian consensus,'' {\em arXiv preprint arXiv:2405.13943}, 2024.

\bibitem{lin2024vastgaussian}
J.~Lin, Z.~Li, X.~Tang, J.~Liu, S.~Liu, J.~Liu, Y.~Lu, X.~Wu, S.~Xu, Y.~Yan, and W.~Yang, ``Vastgaussian: Vast 3d gaussians for large scene reconstruction,'' in {\em CVPR}, 2024.

\bibitem{wang2024pygs}
Z.~Wang and D.~Xu, ``Pygs: Large-scale scene representation with pyramidal 3d gaussian splatting,'' {\em arXiv preprint arXiv:2405.16829}, 2024.

\bibitem{liu2025occlugaussian}
S.~Liu, X.~Tang, Z.~Li, Y.~He, C.~Ye, J.~Liu, B.~Huang, S.~Zhou, and X.~Wu, ``Occlugaussian: Occlusion-aware gaussian splatting for large scene reconstruction and rendering,'' {\em arXiv preprint arXiv:2503.16177}, 2025.

\bibitem{jacobs1991adaptive}
R.~A. Jacobs, M.~I. Jordan, S.~J. Nowlan, and G.~E. Hinton, ``Adaptive mixtures of local experts,'' {\em Neural computation}, vol.~3, no.~1, pp.~79--87, 1991.

\bibitem{jordan1994hierarchical}
M.~I. Jordan and R.~A. Jacobs, ``Hierarchical mixtures of experts and the em algorithm,'' {\em Neural computation}, vol.~6, no.~2, pp.~181--214, 1994.

\bibitem{DBLP:conf/iclr/LepikhinLXCFHKS21}
D.~Lepikhin, H.~Lee, Y.~Xu, D.~Chen, O.~Firat, Y.~Huang, M.~Krikun, N.~Shazeer, and Z.~Chen, ``Gshard: Scaling giant models with conditional computation and automatic sharding,'' in {\em ICLR}, 2021.

\bibitem{DBLP:conf/nips/RiquelmePMNJPKH21}
C.~Riquelme, J.~Puigcerver, B.~Mustafa, M.~Neumann, R.~Jenatton, A.~S. Pinto, D.~Keysers, and N.~Houlsby, ``Scaling vision with sparse mixture of experts,'' in {\em NeurIPS}, 2021.

\bibitem{ye2023taskexpert}
H.~Ye and D.~Xu, ``Taskexpert: Dynamically assembling multi-task representations with memorial mixture-of-experts,'' in {\em ICCV}, 2023.

\bibitem{rogozhnikov2021einops}
A.~Rogozhnikov, ``Einops: Clear and reliable tensor manipulations with einstein-like notation,'' in {\em ICLR}, 2021.

\bibitem{roller2021hash}
S.~Roller, S.~Sukhbaatar, J.~Weston, {\em et~al.}, ``Hash layers for large sparse models,'' {\em NeurIPS}, 2021.

\bibitem{lewis2021base}
M.~Lewis, S.~Bhosale, T.~Dettmers, N.~Goyal, and L.~Zettlemoyer, ``Base layers: Simplifying training of large, sparse models,'' in {\em ICML}, pp.~6265--6274, PMLR, 2021.

\bibitem{zhou2022mixture}
Y.~Zhou, T.~Lei, H.~Liu, N.~Du, Y.~Huang, V.~Zhao, A.~M. Dai, Q.~V. Le, J.~Laudon, {\em et~al.}, ``Mixture-of-experts with expert choice routing,'' {\em Advances in Neural Information Processing Systems}, vol.~35, pp.~7103--7114, 2022.

\bibitem{shazeer2018mesh}
N.~Shazeer, Y.~Cheng, N.~Parmar, D.~Tran, A.~Vaswani, P.~Koanantakool, P.~Hawkins, H.~Lee, M.~Hong, C.~Young, {\em et~al.}, ``Mesh-tensorflow: Deep learning for supercomputers,'' {\em NeurIPS}, 2018.

\bibitem{DBLP:conf/icml/RajbhandariLYZA22}
S.~Rajbhandari, C.~Li, Z.~Yao, M.~Zhang, R.~Y. Aminabadi, A.~A. Awan, J.~Rasley, and Y.~He, ``Deepspeed-moe: Advancing mixture-of-experts inference and training to power next-generation {AI} scale,'' in {\em ICML}, 2022.

\bibitem{hwang2022tutel}
C.~Hwang, W.~Cui, Y.~Xiong, Z.~Yang, Z.~Liu, H.~Hu, Z.~Wang, R.~Salas, J.~Jose, P.~Ram, {\em et~al.}, ``Tutel: Adaptive mixture-of-experts at scale,'' {\em arXiv preprint arXiv:2206.03382}, 2022.

\bibitem{Barron_2022_CVPR}
J.~T. Barron, B.~Mildenhall, D.~Verbin, P.~P. Srinivasan, and P.~Hedman, ``Mip-nerf 360: Unbounded anti-aliased neural radiance fields,'' in {\em CVPR}, 2022.

\bibitem{liu2024deepseek}
A.~Liu, B.~Feng, B.~Xue, B.~Wang, B.~Wu, C.~Lu, C.~Zhao, C.~Deng, C.~Zhang, C.~Ruan, {\em et~al.}, ``Deepseek-v3 technical report,'' {\em arXiv preprint arXiv:2412.19437}, 2024.

\bibitem{li2023matrixcity}
Y.~Li, L.~Jiang, L.~Xu, Y.~Xiangli, Z.~Wang, D.~Lin, and B.~Dai, ``Matrixcity: A large-scale city dataset for city-scale neural rendering and beyond,'' {\em arXiv e-prints}, pp.~arXiv--2308, 2023.

\bibitem{wang2004image}
Z.~Wang, A.~C. Bovik, H.~R. Sheikh, and E.~P. Simoncelli, ``Image quality assessment: from error visibility to structural similarity,'' {\em TIP}, 2004.

\bibitem{zhang2018unreasonable}
R.~Zhang, P.~Isola, A.~A. Efros, E.~Shechtman, and O.~Wang, ``The unreasonable effectiveness of deep features as a perceptual metric,'' in {\em CVPR}, 2018.

\bibitem{rusu20113d}
R.~B. Rusu and S.~Cousins, ``3d is here: Point cloud library (pcl),'' in {\em ICRA}, IEEE, 2011.

\bibitem{barron2023zip}
J.~T. Barron, B.~Mildenhall, D.~Verbin, P.~P. Srinivasan, and P.~Hedman, ``Zip-nerf: Anti-aliased grid-based neural radiance fields,'' in {\em ICCV}, pp.~19697--19705, 2023.

\end{thebibliography}

\begin{IEEEbiography}[{\includegraphics[width=1in,height=1.25in,clip,keepaspectratio]{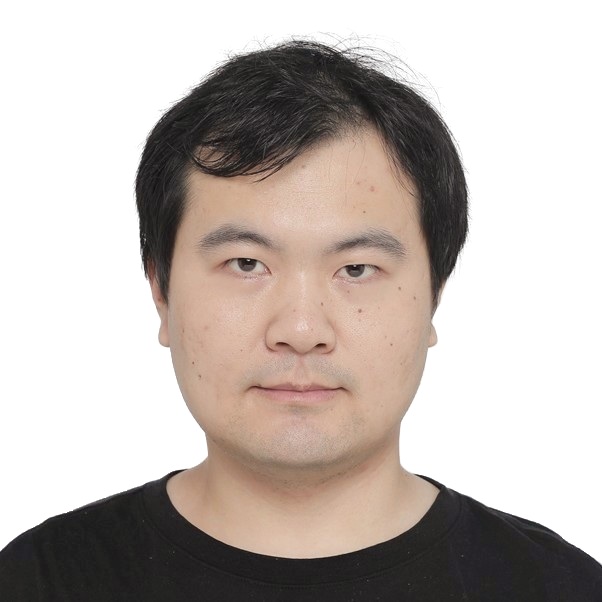}}]{Zhenxing Mi} is a Ph.D. student in the Department of Computer Science and Engineering at Hong Kong University of Science and Technology. He received his M.E. and B.E. from Huazhong University of Science and Technology. His research interest lies in deep learning for large-scale 3D reconstruction and novel view synthesis.
\end{IEEEbiography}
\begin{IEEEbiography}[{\includegraphics[width=1in,height=1.25in,clip,keepaspectratio]{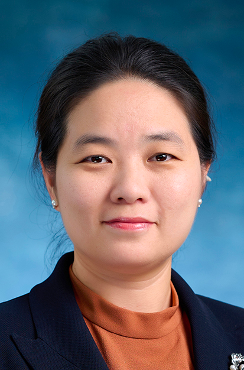}}]{Ping Yin} holds a Ph.D. in Control Theory and Control Engineering from University of Science and Technology Beijing, a Postdoctoral Fellow of Shandong University, and a Senior Engineer. She is an expert of Shandong Provincial Talent Project and a Haiyou Plan Industry Leading Talent. She has led or contributed to 5 provincial and ministerial projects, published 9 journal or conference papers, authored 1 monograph, holds 24 patents, contributed to 4 standards, and co-compiled an industrial white paper.
\end{IEEEbiography}
\begin{IEEEbiography}[{\includegraphics[width=1in,height=1.25in,clip,keepaspectratio]{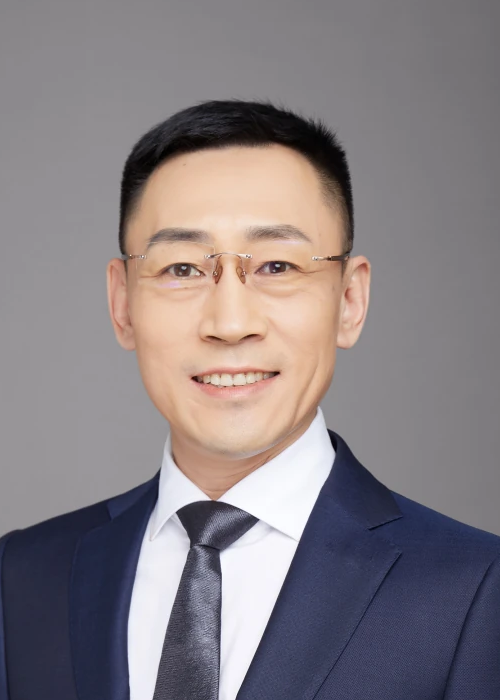}}]{Xue Xiao} is the Chief Engineer of Inspur Group, leading technological innovation in cloud computing and AI at Inspur. He is an expert of Shandong Provincial Talent Project, the Chief Engineer of the China Knowledge Center for Engineering Science and Technology of the Chinese Academy of Engineering, a member of the National Technical Committee 609 on Data of Standardization Administration of China, and an industrial director of the Inspur School of Artificial Intelligence at Shandong University.
\end{IEEEbiography}
\begin{IEEEbiography}[{\includegraphics[width=1in,height=1.25in,clip,keepaspectratio]{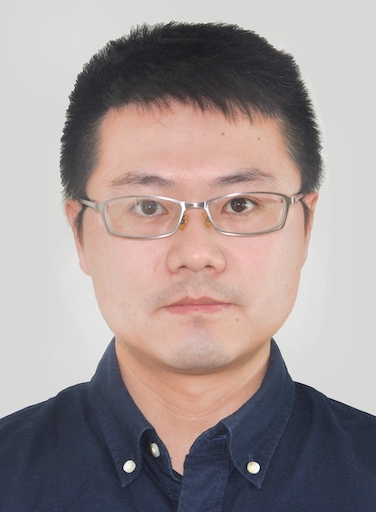}}]{Dan Xu} is an Assistant Professor in the Department of Computer Science and Engineering at HKUST. He was a Postdoctoral Research Fellow in VGG at the University of Oxford. He was a Ph.D. in the Department of Computer Science at
the University of Trento. He was also a student research
assistant at MM Lab at the Chinese University of
Hong Kong. He received the best scientific paper
award at ICPR 2016, and a Best Paper Nominee
at ACM MM 2018. He served as Area Chair at multiple main-stream conferences including NeurIPS, ICML, ICLR, CVPR, ICCV, ECCV, AAAI, IJCAI, ACM Multimedia, and WACV.
\end{IEEEbiography}

\end{document}